\documentclass{article}

\usepackage{times}
\usepackage{graphicx} % more modern
\usepackage{subfigure} 

\usepackage{natbib}

\usepackage{algorithm}
\usepackage{algorithmic}

\usepackage{hyperref}

\usepackage[accepted]{icml2015}

\icmltitlerunning{Cheap Bandits}

\usepackage{amsmath}
\usepackage{amssymb,amsbsy,epsfig,float}

\newtheorem{thm}{Theorem}
\newtheorem{lemma}{Lemma}
\newtheorem{proposition}{Proposition}
\newtheorem{corol}{Corollary}

\newtheorem{remark}{Remark}
\newtheorem{defn}{Definition}

\newcommand{\BS}{\boldsymbol}
\newcommand{\GFT}{\tilde{\boldsymbol{s}}}
\newcommand{\U}{\boldsymbol{\alpha}^*}

\newcommand{\norm}[2]{\|\boldsymbol{#1}\|_{\boldsymbol{#2}}}
\newcommand{\D}{\kappa}

\begin{document} 
	
	\twocolumn[
	\icmltitle{Cheap Bandits}
	
	% unless you've provided the [accepted] option to the icml2015
	\icmlauthor{Manjesh Kumar Hanawal}{mhanawal@bu.edu}
	\icmladdress{Department of ECE, Boston University, Boston, Massachusetts, 02215 USA}
	\icmlauthor{Venkatesh Saligrama}{srv@bu.edu}
    \icmladdress{Department of ECE, Boston University, Boston, Massachusetts, 02215 USA}
    \icmlauthor{Michal Valko}{michal.valko@inria.fr}
 \icmladdress{INRIA Lille - Nord Europe, SequeL team, 40 avenue Halley 59650, Villeneuve d'Ascq, France}
\icmlauthor{R\' emi Munos}{remi.munos@inria.fr}
 \icmladdress{INRIA Lille - Nord Europe, SequeL team, France and Google DeepMind, United Kingdom}
	
	\icmlkeywords{Spectral Bandits}
	
	\vskip 0.3in
	]

\begin{abstract}
We consider stochastic sequential learning problems where the learner can 
observe the \textit{average reward of several actions}. Such a setting is 
interesting in many applications involving monitoring and surveillance, where the set of the actions 
to observe represent some (geographical) area. The importance of this setting is 
that in these applications, it is actually \textit{cheaper} to observe 
average reward of a group of actions rather than the reward of a single action.
We show that when the reward is \textit{smooth} over a given graph representing 
the neighboring actions, we can maximize the cumulative reward of learning
while \textit{minimizing the sensing cost}. In this paper we propose CheapUCB, 
an algorithm that matches the regret guarantees of the known algorithms for 
this setting and at the same time guarantees a linear cost again over them. As a by-product of our analysis, we establish a $\Omega(\sqrt{dT})$ lower bound on the cumulative regret of spectral bandits for a class of graphs with effective dimension $d$.
\end{abstract}

\section{Introduction}
\label{sec:Intro}
% Efficient representation and analysis of large scale data is possible 
% exploiting their structural properties. 

In many online learning and \textit{bandit problems}, the learner is asked to 
select a \textit{single action} for which it obtains a (possibly contextual) feedback.
However, in many scenarios such as surveillance, monitoring and exploration of
a large area or network, it is often cheaper to obtain an average reward for a {\it group of actions} 
rather than a reward for a single one. 
In this paper, we therefore study \textit{group actions} 
and formalize this setting as 
\textit{cheap bandits} on graph structured data. Nodes and edges in our graph
model the geometric structure of the data and we associate signals (rewards) with each node. 
We are interested in problems where the actions are a \emph{collection of nodes}. Our objective is to locate
nodes with largest rewards.

The cost-aspect of our problem arises in \emph{sensor networks }(SNETs) for target 
localization and identification. In SNETs sensors have limited sensing 
range~\cite{ermis10, IPSN2005_AdaptiveStatisticalSampling_ErmisSaligrama}and can reliably sense/identify targets only in their 
vicinity. To conserve battery power, sleep/awake scheduling is 
used~\cite{vvv_sleep,TSP2008_EfficientSensorManagement_AeronSaligCasta}, wherein a {\it group of sensors} is  woken up 
sequentially based on probable locations of target. The {\it group of sensors} 
minimize transmit energy through coherent beamforming of sensed signal, which is 
then received as an average reward/signal at the receiver. While coherent beam 
forming is cheaper, it nevertheless increases target ambiguity since the sensed 
field degrades with distance from target.
%
%Beamforming with set of sensors requires lower energy expenditure over a single sensor and is hence cheaper but comes at a cost of more target ambiguity. 
A similar scenario arises in aerial reconnaissance as well: Larger areas can be surveilled at higher altitudes more
quickly (cheaper) but at the cost of more target ambiguity. 

Moreover, sensing average rewards through group actions, in the initial phases, is also meaningful. Rewards in many applications are typically {\em smooth band-limited graph signals}~\cite{ICASSP2013_SignalProcessingTechniques_NarGadOrt} with the sensing field decaying smoothly with distance from the target. In addition to SNETs 
\cite{ICASSP2012_GraphSpectralCompressed_ZhuRabbat}, smooth graph signals also arise in \emph{social networks} 
\cite{NAS2002_CommunityStructureinSocial_GirvanNewman}, and \emph{recommender systems}. 
Signals on graphs is an emerging area in \emph{signal processing} (SP) but the emphasis 
is on reconstruction through sampling and interpolation from a small subset of 
nodes~\cite{SPMagazine2013_TheEmergingField_ShuNarOrtVander}. In contrast, our 
goal is in locating the maxima of graph signals rather than reconstruction. 
Nevertheless, SP does provide us with the key insight that whenever the graph 
signal is smooth, 
we can obtain information about a location by sampling its neighborhood.

Our approach is to sequentially discover the nodes with optimal reward. We model this problem as an instance of {\em linear 
bandits} \cite{auer2002using, COLT08_StochasticLinearOptimization_DaniHayesKakad,
WWW10_Contextaulbandits_LiChuWei} that links the reward of nodes through an 
unknown parameter. A bandit setting for smooth signals was recently studied 
by~\citet{valko2014spectral}, however \emph{neglecting the signal 
cost}. While typically bandit algorithms aim to minimize the regret, we aim 
to minimize \emph{both regret and the signal cost}. Nevertheless, we do not want to tradeoff  the regret for cost. In particular, 
we are not compromising regret for cost, neither we seek a Pareto frontier of two objectives. We seek algorithms that minimize the cost of sensing and at the same time attain, the state-of-the-art regret guarantees. 

Notice that our setting directly generalizes the traditional setting with 
single action per time step as the arms themselves are graph 
signals. We define cost of each arm in terms of their {\em graph Fourier 
transform}. The cost is quadratic in nature and  assigns higher cost to arms 
that collect average information from a smaller set of neighbors. Our goal is to 
collect higher reward from the nodes while keeping the total cost small. 
However, there is a tradeoff in choosing low cost signals and higher 
reward collection: The arms collecting reward from individual nodes cost 
more, but give more specific information about node's reward and hence provide 
better estimates. On other hand, arms that collect 
average reward from subset of its neighbors cost less, but only give crude estimate 
of the reward function. In this paper, we develop an algorithm maximizing the  
reward collection while keeping the cost low.

\section{Related Work}
\label{sec:RelatedWork}
There are several other bandit and online learning settings that consider 
costs~\cite{tran-thang2012knapsack,badanidiyuru2013bandits,ding2013multi-armed,
COLT14_ResourcefulContextual_BandaniLangSlivk,zolghadr2013online,cesa-bianchi2013online}. The 
first set is referred to as \emph{budgeted 
bandits}~\cite{tran-thang2012knapsack} or \emph{bandits with 
knapsacks}~\cite{badanidiyuru2013bandits}, where each single arm is associated 
with a cost. This cost can be known or unknown~\cite{ding2013multi-armed} and 
can depend on a given context~\cite{COLT14_ResourcefulContextual_BandaniLangSlivk}.
The goal there is in general to minimize the regret as a function of budget 
instead of time or to minimize regret under budget constraints, where there is 
no advantage in not spending all the budget. 
Our goal is different as we care both about minimizing the budget and 
minimizing the regret as a function of time.
Another cost setting considers cost for observing features from which the 
learner can build its prediction~\cite{zolghadr2013online}. This is different 
from our consideration of cost, which is inversely proportional to the sensing 
area. Finally, in the 
adversarial setting~\cite{cesa-bianchi2013online}, considers cost for switching 
actions.

The most related graph bandits setting to ours is
by~\citet{valko2014spectral}
on which we build this paper. Another graph bandit setting 
considers side information, when the learner obtains besides the reward of the 
node it chooses, also the rewards of the neighbors~\cite{mannor2011from,alon2013from,caron2012leveraging,kocak2014efficient}. Finally 
a different graph bandit setup is gang of (multiple) bandits considered 
in~\cite{cesa-bianchi2013gang} and online clustering of bandits 
in~\cite{gentile2014online}.

Our main contribution is the incorporation of \textit{sensing cost} 
into learning in linear bandit problems while simultaneously minimizing two 
performance metrics: cumulative regret and the cumulative sensing cost. 
We develop CheapUCB, the algorithm that guarantees regret bound of the order 
$d\sqrt{T}$, where $d$ is the \emph{effective dimension} and $T$ is the number of 
rounds. This regret bound is of the same order as  
SpectralUCB~\cite{valko2014spectral} that does not take 
cost into consideration. However, we show that our algorithm provides a cost 
saving that is linear in $T$ compared to the cost of SpectralUCB.
The effective dimension $d$ that appears in the bound is a dimension typically 
smaller in real-world graphs as compared to number of nodes $N$. 
This is in contrast with linear bandits that can achieve in this graph 
setting the regret of $N\sqrt{T}$ or $\sqrt{NT}$. However, our ideas of cheap 
sensing are directly applicable to the linear bandit setting as well. As a by-product of our analysis, we establish a $\Omega(\sqrt{dT})$ lower bound on the cumulative regret for a class of graphs with effective dimension $d$.

\section{Problem Setup}
\label{Sec:Settings}
Let $G=(\mathcal{V},\mathcal{E})$ denote an undirected graph with number of nodes 
$|\mathcal{V}|=N$. We assume that degree of all the nodes is bounded by $\D$. Let $\mathbf{s}: \mathcal{V} \rightarrow \mathcal{R}$ denote a signal on 
$G$, and $\mathcal{S}$ the set of all possible signals on 
$\mathcal{G}$. Let $\BS{L}=\BS{D}-\BS{A}$ denote the unnormalized Laplacian of the 
graph $G$, where $\BS{A}=\{a_{ij}\}$ is the adjacency matrix and $\BS{D}$ 
is the diagonal matrix with $\BS{D}_{ii}=\sum_{j} {a_{ij}}$. We emphasize that our main results extend to weighted graphs if we replace the matrix $\BS{A}$ with the edge weight matrix $\BS{W}$. We work with matrix $\BS{A}$ for simplicity of exposition.  We denote the 
eigenvalues of $\BS{L}$ as $0=\lambda_1 \leq 
\lambda_2 \leq \cdots \leq \lambda_N$, and the 
corresponding eigenvectors as $ \bf q_1, q_2, \cdots, q_N$. Equivalently, we 
write $\BS{L}=\BS{Q}  \Lambda_\mathcal{L}  Q^\prime$, where ${\bf 
\Lambda}_\mathcal{L}=diag(\lambda_1,\lambda_2, \cdots, 
\lambda_N)$ and $\mathbf{Q}$ is the $N\times N$ orthonormal matrix with 
eigenvectors in columns. We denote transpose of $\mathbf{a}$ as 
$\mathbf{a}^\prime$, and all vectors are by default column vectors. For a given matrix $\BS{V}$, we denote $\BS{V}$-norm of a vector $\mathbf{a}$ as $\| 
\mathbf{a} \|_V=\sqrt{\mathbf{a}^\prime \BS{V}	 \mathbf{a}}$.

\subsection{Reward function}
\label{subsec:RewardFunction}
We define a reward function on a graph $G$ as a linear combination of the eigenvectors. For a given parameter vector $ \boldsymbol{\alpha} \in \mathcal{R}^N$, let $\BS{f}_{\BS{\alpha}}: \mathcal{V}\rightarrow \mathcal{R}$ denote the reward function on the nodes defined as 
\[\BS{f}_{ \BS{\alpha}}=  \BS{Q} \BS{\alpha}.\]
The parameter $\BS{\alpha}$ can be suitably penalized to control the smoothness of the reward function. For instance, 
if we choose $\BS{\alpha}$ such that large coefficients correspond to the 
eigenvectors associated with small eigenvalues then 
$f_{\BS{\alpha}}$ is a smooth function of $G$ 
\cite{JMLR2008_ManifoldRegularization_BelkinNiyogiSindh}.
We denote the \emph{unknown} parameter that defines the true reward function as $\boldsymbol{\alpha}^*$. We denote the reward of node $i$ as $\BS{f}_{\U}(i)$.

In our setting, the arms are nodes \emph{and} the subsets of their neighbors. When an arm is selected, we observe only the average of the rewards of the nodes selected by that arm. To make this notion formal, we associate arms with \emph{probe signals} on graphs.  

\subsection{Probes}
\label{subsec:ActionSet}
%In the literature on sampling theory in graphs, a functions 
%defined on the nodes is referred to as signal 
%\cite{SPMagazine2013_TheEmergingField_ShuNarOrtVander}. Deviating from this convention, we refer to $f_{\boldsymbol{\alpha}^*}$ as reward function and other 
%functions defined as weights on the nodes as probe-signals. 
Let $\mathcal{S}\subseteq 
\left \{\mathbf{s} \in [0,1]^N : \sum_{i=1}^{N}s_i=1\right \}$ denote the set of probes. We use the word probe and action interchangeably. A probe is a signal with its width corresponding to the support of the signal $s$. For instance, it could correspond to the region-of-coverage or region-of-interest probed by a radar pulse. 
Thus each $\mathbf{s} \in \mathcal{S}$ is of the form 
$s_i=1/\text{supp}(\mathbf{s})$, for all $i=1,2,\cdots,N$, where 
$\text{supp}(\mathbf{s})$ denotes the number of positive elements in 
$\mathbf{s}$. The inner product of $\BS{f}_{\BS{\alpha}^*}$ and a probe 
$\mathbf{s}$ is the average reward of  $\text{supp}(\mathbf{s})$ number of nodes. 

We parametrize a probe in terms of its width $w \in [N]$ and let the set of probes of width $w$ to be $\tilde{\mathcal{S}}_w=\{\mathbf{s}\in \mathcal{S}: 
\text{supp}(\mathbf{s})=w\}$. For a given 
$w> 0$, our focus in this paper is on probes with uniformly weighted components, which are limited to neighborhoods of each node on the graph. We denote the collection of these probes as  $\mathcal{S}_w \subset 
\tilde{\mathcal{S}}_w$, which has $N$ elements. We denote the element in $\mathcal{S}_w$ associated with node $i$ as 
$\boldsymbol{s}_i^w$. Suppose node $i$ has neighbors at $\{j_1, j_2, \cdots 
j_{w-1}\}$, then $\boldsymbol{s}_i^w$ is described as:
\begin{equation}
s_{ik}^w=\begin{cases}
1/w \; & \mbox{if $k=i$}  \\
1/w \;\; &\mbox{if}  \;\;k=j_i,\;\; i=1,2,\cdots, w-1\\
0 & \mbox{otherwise.}
\end{cases}
\end{equation}
If node $i$ has more than $w$ neighbors, there can be multiple ways to define $\mathbf{s}_i^w$ depending on the choice of its neighbors. When $w$ is less than degree of node $i$, in defining $\mathbf{s}_i^w$ we only consider neighbors with larger edge weights. If all the weights are the same, then we select $w$ neighbors arbitrarily. Note that $|\mathcal{S}_w|=N$ for all $w$. In the following we write `probing with $\mathbf{s}$' to mean that $\mathbf{s}$ is used to get information from nodes of graph $G$. 

We define the arms as the set $$\mathcal{S}_D:=\{\mathcal{S}_w: 
w=1,2,\cdots,N\}.$$ Compared to multi-arm and linear bandits, the number 
of arms $K$ is $\mathcal{O}(N^2)$ and the contexts have dimension $N$.

\subsection{Cost of probes}
\label{subsec:Cost}
 The cost of the arms are defined using the spectral properties of their associated graph probes. Let $\tilde{\mathbf{s}}$ denote the {\em graph Fourier transform} (GFT) of probe $\mathbf{s} \in \mathcal{S}$. Analogous to Fourier transform of a continuous function, GFT gives amplitudes associated with graph frequencies. The GFT coefficient of a probe on frequency $\lambda_i, i=1,2 \cdots,N$ is obtained by projecting it on $\mathbf{q}_i$, i.e.,
\[\bf \tilde s= Q^\prime s,\]
where $\GFT_i, i=1,2,\cdots,N$ is the GFT coefficient associated with frequency $\lambda_i$.
%We define cost of  probing the graph with a signal as sum of square of its GFT coefficients weighted by the corresponding frequency. 
%
Let $C: \mathcal{S} \rightarrow \mathcal{R}_+$ denote the cost function.
% \footnote{In defining the cost we set $W=A$, where $A$ is the adjacency matrix. 
% Also, we added self loops (of edge weights 1) to the nodes and made the graph 
% symmetric with the degree of all nodes equal to $N$.}. 
Then the cost of the 
probe $\mathbf{s}$ is described by
\[C(\mathbf{s})=\sum_{i \sim j}(s_i-s_j)^2,\]
where the summation is over all the unordered node pairs $\{i,j\}$ for which node $i$ is adjacent to node $j$. We motivate this cost function from the SNET perspective where probes with large width are relatively cheap. We first observe that the cost of a constant probe is zero. For a probe, $\mathbf{s}_i^w \in \mathcal{S}_w$, of width $w$ it follows that\footnote{We symmetrized the graph by adding self loops to all the nodes to make their degree (number of neighbors) $N$, and normalized the cost by $N$. },
\begin{equation}
\label{eqn:SignalCost}
C(\mathbf{s}_i^w)= \frac{w-1}{w^2}\left( 1- \frac{1}{N}\right) + \frac{1}{w^2}.
\end{equation}
Note that the cost of $w$- width probe associated with node $i$ depends only on 
its width $w$. For $w=1$, $C(\mathbf{s}_i^1)=1$ for all $i=1,2,\cdots, N$. That is, the 
cost of probing individual nodes of the graph is the same. Also note that  
$C(\mathbf{s}_i^w)$ is decreasing in $w$, implying that probing a node is more  
costly than probing a subset of its neighbors. 

Alternatively, we can associate probe costs with eigenvalues of the graph Laplacian. Constant probes corresponds to the zero eigenvalue of the graph Laplacian. More generally, we see that,
\[C(\mathbf{s})=\sum_{i \sim j}(s_i-s_j)^2 = \mathbf{s}^\prime \mathcal{L} \mathbf{s} = \sum_{i=1}^N \lambda_i \tilde{s}_i^2=  {\bf \tilde s^\prime \Lambda_{\mathcal{L}} \tilde s}.\]
It follows that $C(\mathbf{s})=\| \mathbf{s} \|^2_{\mathcal{L}}$. 
The operation of pulling an arm and observing a reward is equivalent to probing the graph with a probe. This results in a value that is the inner product of the probe signal and graph reward function. We write the reward in the probe space $\mathcal{S}_D$ as follows. Let $ F_{G}: \mathcal{S} \rightarrow \mathcal{R}$ defined as 
\[F_{G}(\mathbf{s})=\mathbf{s}^\prime\boldsymbol{Q} \boldsymbol{\alpha}^*=\mathbf{\tilde s}^\prime \boldsymbol{\alpha}^*\]
denote the reward obtained from probe $\mathbf{s}$.
Thus, each arm gives a reward that is linear, and has quadratic cost, in its GFT coefficients. In terms of the linear bandit terminology, the GFT coefficients in $\mathcal{S}_D$ constitute the set of arms.

With the rewards defined in terms of the probes, the optimization of reward function is over the action space. 
Let $\mathbf{s}_*=\arg \max _{\mathbf{s} \in 
\mathcal{S}_D}F_{G}(\mathbf{s})$  denote the probe that gives the  maximum reward. This is a straightforward linear optimization problem if the  function parameter $\boldsymbol{\alpha}^*$ is known. When $\boldsymbol{\alpha}^*$ is unknown we can learn the function through a sequence 
of measurements. 

\subsection{Learning setting and performance metrics}
\label{subsec:LearningSetting}
Our learning setting is the following. The learner uses a policy $\pi : \{1,2, \cdots, T\} \rightarrow \mathcal{S}_D$ that assigns at step $t \leq T$, probe $\pi(t)$. In each step $t$, the recommender incurs a cost $C(\pi(t))$ and obtains a noisy reward such that
\[r_t = F_{G}(\pi(t)) + \varepsilon_t,\]
where $\varepsilon_t$ is independent $R$-{\em sub Gaussian} for any $t$. 

The cumulative regret of policy $\pi$ is defined as
\begin{eqnarray}
R_T&=& T  F_{G}(\mathbf{s}_*) - \sum_{t=1}^T F_{G}(\pi(t))
\end{eqnarray}
and the total cost incurred up to time $T$ is given by 
\begin{eqnarray}
C_T=\sum_{t=1}^{T}C(\pi(t)).
\end{eqnarray}

The goal of the learner is to learn a policy $\pi$ that minimizes total cost $C_T$ while keeping the
cumulative (pseudo) regret $R_T$ as low as possible.

{\bf Node vs.~Group actions:} The set $\mathcal{S}_D$ allows actions that can probe a node (node-action) or a subset of nodes (group-action). Though the group actions have smaller cost, they only provide average reward information for the selected nodes. In contrast, node actions provide crisper information of the reward for the selected node, but at a cost premium. Thus, an algorithm that uses only node actions can provide a better regret performance compared to the one that takes group actions. But if the algorithms use only node actions, the cumulative cost can be high.

In the following, we first state the regret performance of the SpectralUCB algorithm \cite{valko2014spectral}  that uses only node actions. We then develop an algorithm that aims to achieve the same order of regret using group actions and reducing the total sensing cost.
\vspace{-.2cm}
\section{Node Actions: Spectral Bandits}
\label{sec:SpectralBandit}
If we restrict the action set to $\mathcal{S}_D=\{\mathbf{e}_i: 
i=1,2,\cdots, n\}$, where $\mathbf{e}_i$ denotes a binary vector with $i^{th}$ 
component set to $1$ and all the other components set to $0$, then only node actions are allowed in each step. In this setting, the cost is the same for all the 
actions, i.e., $C(\mathbf{e_i})=1$ for all $i$.

Using these node actions, \citet{valko2014spectral} developed SpectralUCB  that aims to minimize the regret under the assumption that the reward function is smooth.  The smoothness condition is characterized as follows:

\begin{equation}
\label{asm:Smoothness}
\exists \;\; c >0 \;\;\text{such that}\;\; \| \boldsymbol{\alpha^*} 
\|_{\boldsymbol{\Lambda}} \leq c.
\end{equation}

Here ${\bf \Lambda}= \boldsymbol{\Lambda_\mathcal{L}}+ \lambda I$, and $\lambda 
>0$ is used to make $\boldsymbol{\Lambda_\mathcal{L}}$ invertible. The bound $c$ 
characterizes the smoothness of the reward. When $c$ is small, the 
rewards on the neighboring nodes are more similar. In particular, when the 
reward function is a constant, then  $c=0$.
To characterize the regret performance of  SpectralUCB, \citet{valko2014spectral} introduced the notion of {\em effective dimension} defined as follows:

\begin{defn}[Effective dimension]
	\label{dfn:EffectiveDimension}
	For graph G, let us denote $\lambda=\lambda_1\leq \lambda_2 \cdots \leq \lambda_N$  the diagonal elements of $\boldsymbol{\Lambda}$. Given $T$, effective dimension is the largest $d$ such that:
	\begin{equation}\label{eqn:EffectiveDimension}
		(d-1)\lambda_d \leq \frac{T}{\log(T/\lambda +1)} < d \lambda_{d+1}.
 \end{equation}
\end{defn}

\begin{thm}\cite{valko2014spectral}
\label{thm:Regret1}
 The cumulative regret of SpectralUCB is bounded with probability at least $1-\delta$ as:
	\begin{eqnarray*}
		R_T \leq \left (8R\sqrt{d\log(1+ T/\lambda)+2 \log (1/\delta)}+4c\right )\\
	\hspace{-1cm}	\times\sqrt{dT\log(1+ T/\lambda)},
	\end{eqnarray*}
\end{thm}

\begin{lemma}
The total cost of the SpectralUCB is $C_T=T$.
\end{lemma}

Note that effective dimension depends on $T$ and also on how fast the eigenvalues grow. The regret performance of SpectralUCB is good when $d$ is small, which occurs when the eigenspectrum exhibits large gaps. For these situations, SpectralUCB performance has a regret that scales as $O(d\sqrt{T})$ for a large range of values of $T$. To see this, notice that in relation (\ref{eqn:EffectiveDimension}) when $\lambda_{d+1}/\lambda_d$ is large,  the value of effective dimension remains unchanged over a large range of $T$ implying that the regret bound of  $O(d\sqrt{T})$ is valid for a large range of values of $T$ with the same $d$. 

There are many graphs for which the effective dimension is small. For example, random graphs are good expanders for which eigenvalues grow fast. Another setting are stochastic block models \cite{NAS2002_CommunityStructureinSocial_GirvanNewman}, that exhibit large eigenvalue gap and are popular in the analysis of social, biological, citation, and information networks.

\section{Group Actions: Cheap Bandits}
\label{sec:CheapBandits}
Recall (Section~\ref{subsec:Cost}) that group actions are cheaper than the node actions. Furthermore, that the cost of group actions is decreasing in group size. In this section, we develop a learning algorithm that aims to minimize the total cost without compromising on the regret using group actions. Specifically, given $T$ and a graph with effective dimension~$d$ our objective is as follows:  
 
\begin{equation}
\label{eqn:Objective}
%\begin{aligned}
\min_{\pi} \, C_T\,\,\,\, \mbox{subject to}\,\,\,\, R_T \lesssim d\sqrt{T}.
%\end{aligned}
\end{equation}
where optimization is over policies defined on the action set $\mathcal{S}_D$ given in subsection \ref{subsec:ActionSet}.

\subsection{Lower bound}
\label{subsec: LowerBound}

The action set used in the above optimization problem is larger than the set used in the SpectralUCB. This raises the question of whether or not the regret order of $d\sqrt{T}$ is too loose particularly when SpectralUCB can realize this bound using a much smaller set of probes.

In this section we derive a $\sqrt{dT}$ lower bound on the expected regret (worst-case) for any algorithm using action space $\mathcal{S}_D$ on graphs with effective dimension~$d$. While this implies that our target in (\ref{eqn:Objective}) should be $\sqrt{dT}$, we follow  \citet{valko2014spectral} and develop a variation of SpectralUCB that obtains the target regret of $d\sqrt{T}$. We leave it as a future work to develop an algorithm that meets the target regret of $\sqrt{dT}$ while minimizing the cost.  

Let $\mathcal{G}_d$ denote a set of graphs with effective dimension $d$. 
For a given policy $\pi, \BS{\alpha}^*,T$ and graph $G$.
Define expected cumulative reward as \[Regret(T, \pi,\U, G)=\mathbb{E}\left [ \sum_{t=1}^T \tilde{\BS{s}}_*\U -   \tilde{\BS{s}}_t\U \Big| \U \right ]\]
where $\tilde{\BS{s}_t}=\pi^\prime(t)\BS{Q}$.

\begin{proposition}
\label{prop: LowerBound}
For any policy $\pi$ and time period $T$, there exists a graph $G \in \mathcal{G}_d$ and a $\U \in \mathcal{R}^d$ representing a smooth reward such that	
\[Regret(T, \pi,\U, G) = \Omega(\sqrt{dT})\]
\end{proposition}

The proof follows by construction of a graph with $d$ disjoint cliques and restricting the rewards to be piecewise constant on the cliques. The problem then reduces to identifying the clique with the highest reward. We then reduce the problem to the multi-arm case, using Theorem~5.1 of \citet{SIAMR02_ANonStochasticMultiArmed_AuerBiachiFreud} and lower bound the minimax risk. See the supplementary material for a detailed proof. 

%We note that regret bound in (\ref{eqn:Objective}) does not mathch the lower bound. Following \cite{valko2014spectral} we first obtain $d\sqrt{T}$ algorithm similar to SpectralUCB. We can also obtain $\sqrt{dT}$  algorithm by modifying SpectralEliminator (see Remark 2).

\subsection{Local smoothness}
In this subsection we show that a smooth reward function on a graph with low effective dimension implies local smoothness of the reward function around each node. Specifically, we establish that the average reward around the neighborhood of a node provides good information about the reward of the node itself. Then, instead of probing a node, we can use group actions to probe its neighborhood and get good estimates of the reward at low cost.

From the discussion in Section \ref{sec:SpectralBandit},  when $d$ is small and there is a large gap between the $\lambda_{d}$ and $\lambda_{d+1}$, SpectralUCB enjoys a small bound on the regret for a large range of values in the interval $[(d-1)\lambda_d,\,d \lambda_{d+1}]$. Intuitively, a large gap between the eigenvalues implies that there is a good partitioning of the graph into tight clusters. Furthermore, the smoothness  assumption implies that the reward of a node and its neighbors within each cluster are similar. 

 Let $\mathcal{N}_i$ denote a set of neighbors of node $i$. The following result provides a relation between the reward of node $i$ and the average reward from $\mathcal{N}_i$ of its neighbors.
 \begin{proposition}
 	\label{prop:Smoothness2}
 	Let $d$ denote the effective dimension and  $\lambda_{d+1}/\lambda_d \geq \mathcal{O}(d^2)$. Let $\U$ satisfy (\ref{asm:Smoothness}). For any node $i$
 	\begin{equation}
 	\label{eqn:LocalSmoothness}
 	\left | \BS{f}_{\U}(i) - \frac{1}{|\mathcal{N}_i|} \sum_{j \in \mathcal{N}_i} \BS{f}_{\U}(j) \right |  \leq c^\prime d /\lambda_{d+1}
 	\end{equation}
 	for all $\mathcal{N}_i$, and $c^\prime= 56\D\sqrt{2\D}c$.
 \end{proposition}
 
The full proof is given in the supplementary material. It is based on $k$-way expansion constant together with bounds on higher order Cheeger inequality~\cite{SODA14_PartioningIntoExpander_GharanTrevisan}. 
Note that (\ref{eqn:LocalSmoothness}) holds for all $i$. However, we only need this to hold for the node with the optimal reward to establish regret performance our algorithm. We rewrite~(\ref{eqn:LocalSmoothness}) for the optimal $i^*$ node using group actions as follows:
\begin{equation}
\label{eqn:LocalSmoothness1}
\left | F_G(\BS{s}_*)- F_G(\BS{s}_*^w) \right | \leq  c^\prime d/\lambda_{d+1} \;\; \mbox{for all} \;\; w\leq  |\mathcal{N}_{i^*}| .
\end{equation}
Though we give the proof of the above result under the technical assumption $\lambda_{d+1}/\lambda_d \geq \mathcal{O}(d^2)$, it holds in cases where eigenvalues grow fast. For example, 
for graphs with strong connectivity property this inequality is trivially satisfied. We can show that $\left | F_G(\mathbf{s}_*) -F_G(\mathbf{s}_*^w)\right |  \leq c/\sqrt{\lambda_2}$ through a standard application of Cauchy-Schwartz inequality. For the model of Barab\' asi-Albert  we get $\lambda_2=\Omega(N^{\gamma})$ with $\gamma > 0$ and for the cliques we get $\lambda_2 = N$.

{\bf General graphs:}
When $\lambda_{d+1}$ is much larger than $\lambda_d$, the above proposition gives a tight relationship between the optimal reward and the average reward from its neighborhood. However, for general graphs this eigenvalue gap assumption is not valid. Motivated by (\ref{eqn:LocalSmoothness1}), we assume that the smooth reward function satisfies the following weaker version for the general graphs. For all $w \leq  |\mathcal{N}_{i^*}|$
\begin{equation}
\label{eqn:LocalSmoothness2}
\left | F_G(\BS{s}_*)- F_G(\BS{s}_*^w) \right | \leq   c^\prime \sqrt{T}w/ \lambda_{d+1}.
\end{equation}
These inequalities get progressively weaker in $T$ and $w$ and can be interpreted as follows. For small values of $T$, we have few rounds for exploration and require stronger assumptions on smoothness. On the other hand, as $T$ increases we have the opportunity to explore and consequently the inequalities are more relaxed. This relaxation of the inequality as a function of the width $w$  characterizes the fact that close neighborhoods around the optimal node provide better information about the optimal reward than a wider neighborhood. 
\subsection{Algorithm: CheapUCB}
\label{Sec: Algorithm}
Below we present an algorithm similar to LinUCB 
\cite{WWW10_Contextaulbandits_LiChuWei} and SpectralUCB 
\cite{valko2014spectral} for regret minimization. 
The main difference between our algorithm and the SpectralUCB algorithm is 
the enlarged action space, which allows for selection of subsets of nodes and associated realization of 
average rewards. Note that when we probe a specific 
node instead of probing a subset of nodes, we get a more 
precise information (though noisy) about the node, but this results in higher 
cost. 

As our goal is to minimize the cost while maintaining a low regret, we handle this requirement by moving sequentially from the least costly probes to expensive ones as we progress. In particular, we split the time horizon into $J$ stages, and as we move from state $j$ to $j+1$ we use  more expensive probes. That means, we use probes with smaller widths as we progress through the different stages of learning. The algorithm uses the probes of different widths in each stage as follows.
Stage $j=1, \dots, J$ consists of time steps from $2^{j-1}$ to $2^{j}-1$ and uses of probes of weight $j$ only.

At each time step $t=1,2,\dots,T,$ we estimate the value of 
$\boldsymbol{\alpha^*}$ by using $\mathit{l}^2$-regularized least square as follows. Let $\{\mathbf{s}_i:=\pi(i), i=1,2,\dots,t\}$ denote the probe selected till time $t$ and $\{r_i, i=1,2,\dots,t\}$ denote the corresponding rewards. The estimate of $\boldsymbol{\alpha^*}$ denoted
$\boldsymbol{\hat\alpha}_t$ is computed as 

\[\hat{\BS{\alpha}}_t= \arg \min _{\BS{\alpha}} \left (\sum_{i=1}^{t} 
\left[\mathbf{s}^\prime_i \boldsymbol{Q \alpha} - r_t \right] ^2 + \| 
\boldsymbol{\alpha} \|^2_{\boldsymbol{\Lambda}} \right ).\]

 \begin{algorithm}[ht]
 \footnotesize
 \caption{CheapUCB}
 \label{Algorithm:CheapBandits}
 \begin{algorithmic}[1]
\STATE \textbf{Input:}
 \STATE $G$: graph
\STATE $T$: number of steps
 \STATE $\lambda, \delta$: regularization and confidence parameters
 \STATE $R,c$: upper bound on noise and norm of $\BS{\alpha}$
\STATE \textbf{Initialization:}
\STATE $d \leftarrow \arg \max\{d: (d-1)\lambda_d \leq T/\log(1+ T/\lambda)\}$
\STATE $\beta \leftarrow 2R \sqrt{d\log(1+T/\lambda) + 2\log( 1/\delta)} + c$
\STATE $\BS{V}_0 \leftarrow \BS{\Lambda}_{L}+\lambda \BS{I}, \BS{S}_0\leftarrow 0, r_0\leftarrow 0$
     \FOR {$j = 1 \to J$}
     \FOR {$t=2^{j-1} \to \min\{2^j-1, T\}$}
     
                   \STATE $\BS{S}_t \leftarrow \BS{S}_{t-1} + r_{t-1} \tilde{\BS{s}}_{t-1}$
                   \STATE $\BS{V}_t \leftarrow \BS{V}_{t-1} + \tilde{\BS{s}}_{t-1} \tilde{\BS{s}}^\prime_{t-1}$ 
                   \STATE $\BS{\hat{\alpha}}_{t} \leftarrow \BS{V}^{-1}_{t} \BS{S}_{t}$ 
                   \STATE $\BS{s}_t \leftarrow \arg \max_{\BS{s} \in \displaystyle\mathcal{S}_{{J -j+1}}} \left (\tilde{\mathbf{s}}^\prime \BS{\hat{\alpha}}_{t}
                   + \beta \norm{\GFT} {V_t^{-1}}  \right )$
                  \ENDFOR 
                  \ENDFOR 
 \end{algorithmic}
  % \vspace{-0.4cm}%
 \end{algorithm}

 \begin{thm}
 \label{thm:Regret1}
 Set $J=\lceil\log T \rceil$ in the algorithm. Let $d$ be the effective dimension and $\lambda$ be the smallest 
eigenvalue  of $\Lambda$. Let $\GFT_t^\prime\U \in [-1, 1]$ for 
all $\mathbf{s} \in \mathcal{S}$, the cumulative 
regret of the algorithm is with probability at least $1-\delta$ bounded as:

(i) If  (\ref{asm:Smoothness})  holds and $\lambda_{d+1}/ \lambda_d \geq \mathcal{O}(d^2) $, then 
 \vspace{-.2cm}
\begin{eqnarray*}
\lefteqn{R_T \leq (8R\sqrt{d\log(1+ T/\lambda)+2 \log (1/\delta)}+4c)}\\
&\times&\hspace{-.3cm}\sqrt{dT\log(1+ T/\lambda)}+ c^\prime d^2 \log_2(T/2)\log(T/\lambda +1),
\end{eqnarray*}

(ii) If (\ref{asm:Smoothness}) and (\ref{eqn:LocalSmoothness2}) hold, then
 \vspace{-.2cm}
 \begin{eqnarray*}
 	\lefteqn{R_T \leq (8R\sqrt{d\log(1+ T/\lambda)+2 \log (1/\delta)}+4c)}\\
 	&\times&\hspace{-.3cm}\sqrt{dT\log(1+ T/\lambda)}+ c^\prime d \sqrt{T/4}\log_2(T/2)\log(T/\lambda +1),
 	\end{eqnarray*}
Moreover, the cumulative cost of CheapUCB is bounded as 
 	\[C_T\leq \sum_{j=1}^{J-1}\frac{2^{j-1}}{J-j+1}\leq \frac{3T}{4}-\frac{1}{2}\]
 \end{thm}

 \begin{remark}
 \label{rmk:Comarison}
Observe that when the eigenvalue gap is large, we get the regret to order $d\sqrt{T}$ within a constant factor satisfying the constraint  (\ref{eqn:Objective}). For the general case,
compared to  SpectralUCB, the regret bound of our algorithm increases 
by an amount of $cd\sqrt{T/2}\log_2(T/2)\log( T/\lambda +1)$, but still it is of 
the order $d\sqrt{T}$.  However, the total cost in CheapUCB is 
smaller than in SpectralUCB by an amount of at least $T/4+1/2$, i.e., cost 
reduction of the order of $T$ is achieved by our algorithm.
 \end{remark}

 \begin{corol}
 	CheapUCB matches the regret performance of SpectralUCB and provides a cost gain of $\mathcal{O}(T)$.
 \end{corol}

    \begin{figure*}[!t]
    	\centering
    	\vspace{-.3cm}
    	\subfigure[Regret for BA graph ]{\includegraphics[scale=0.2]{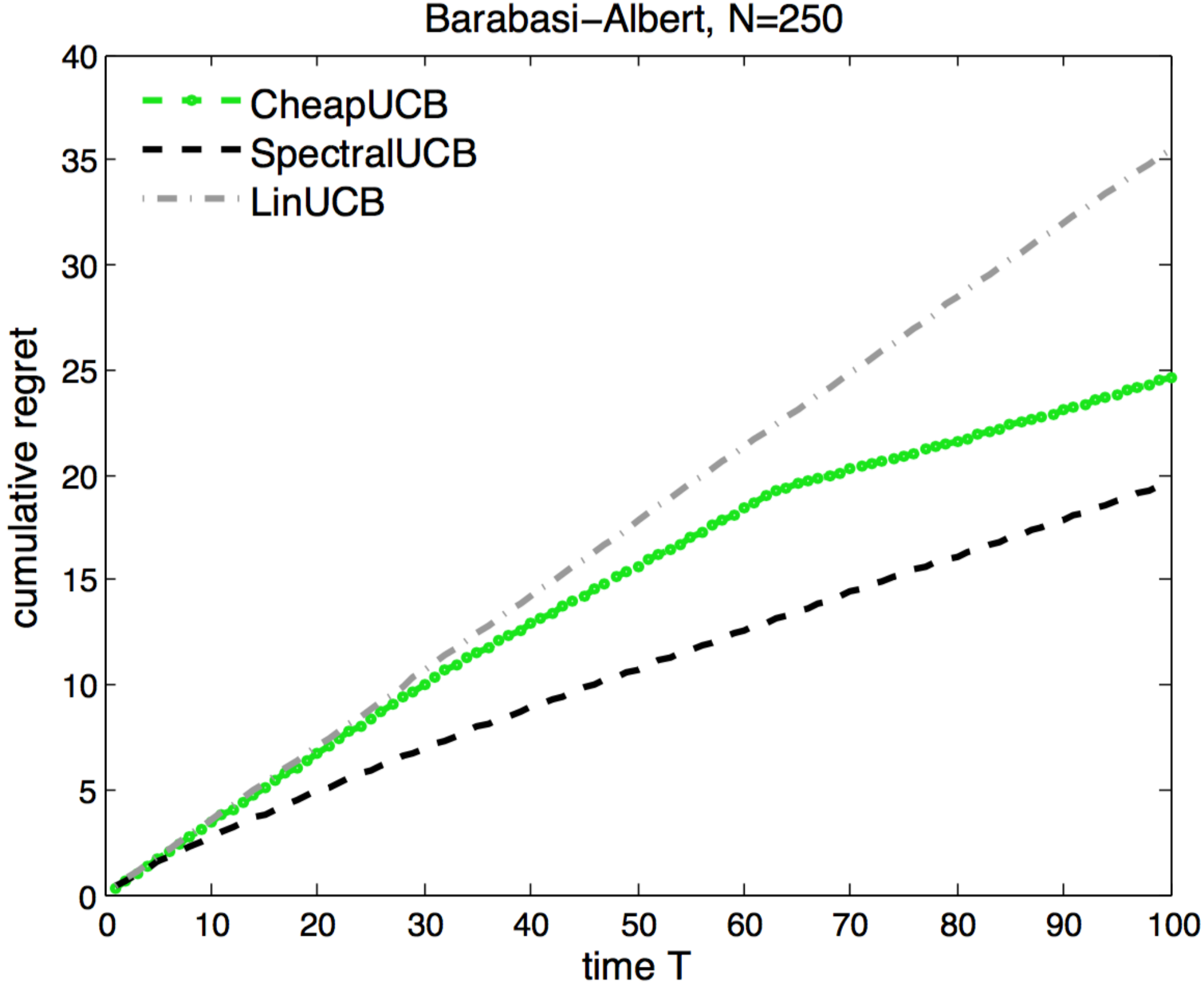}}
    	\subfigure[Cost for BA graph]{\includegraphics[scale=0.2]{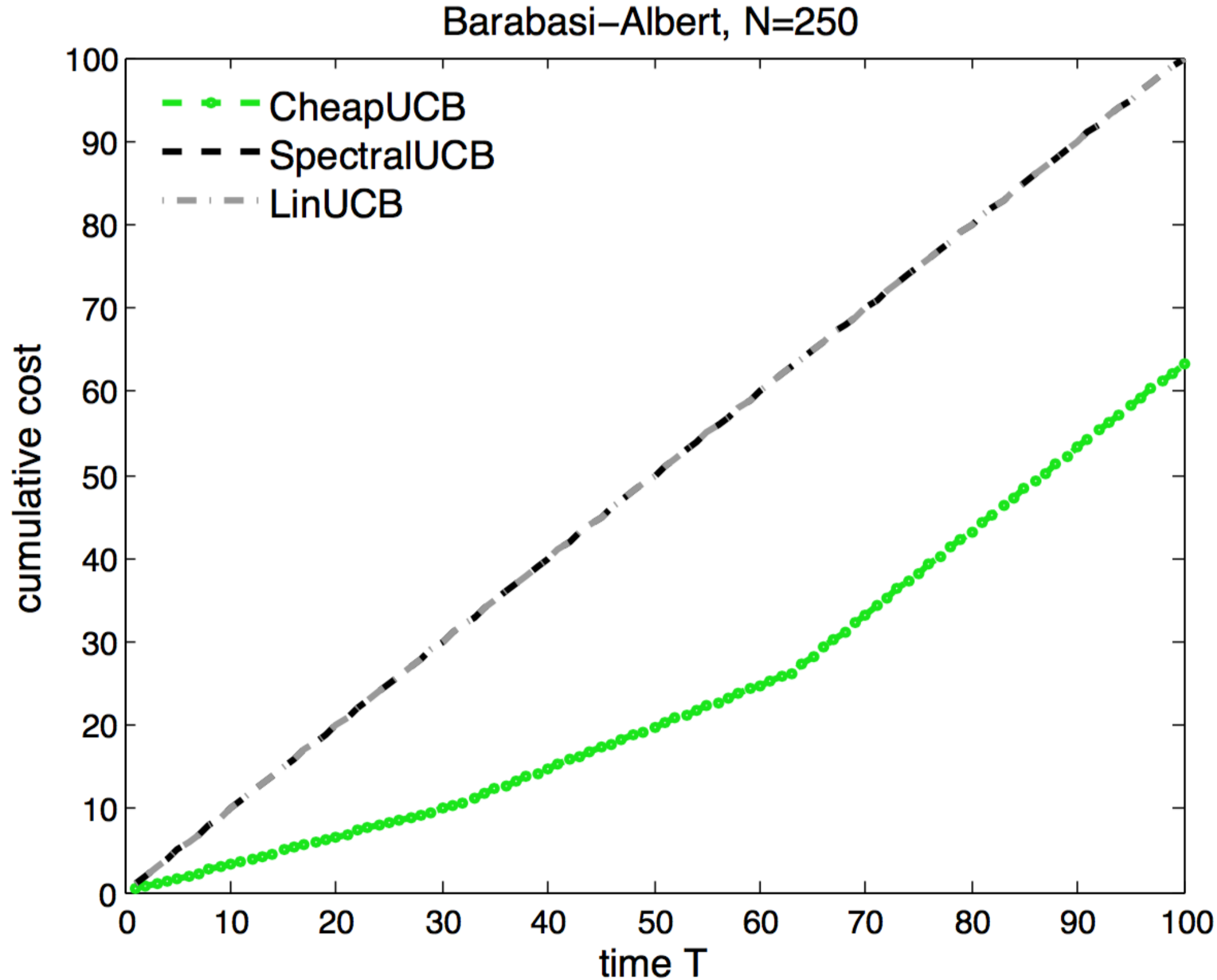}}
    	\subfigure[Regret for ER graph ]{\includegraphics[scale=0.2]{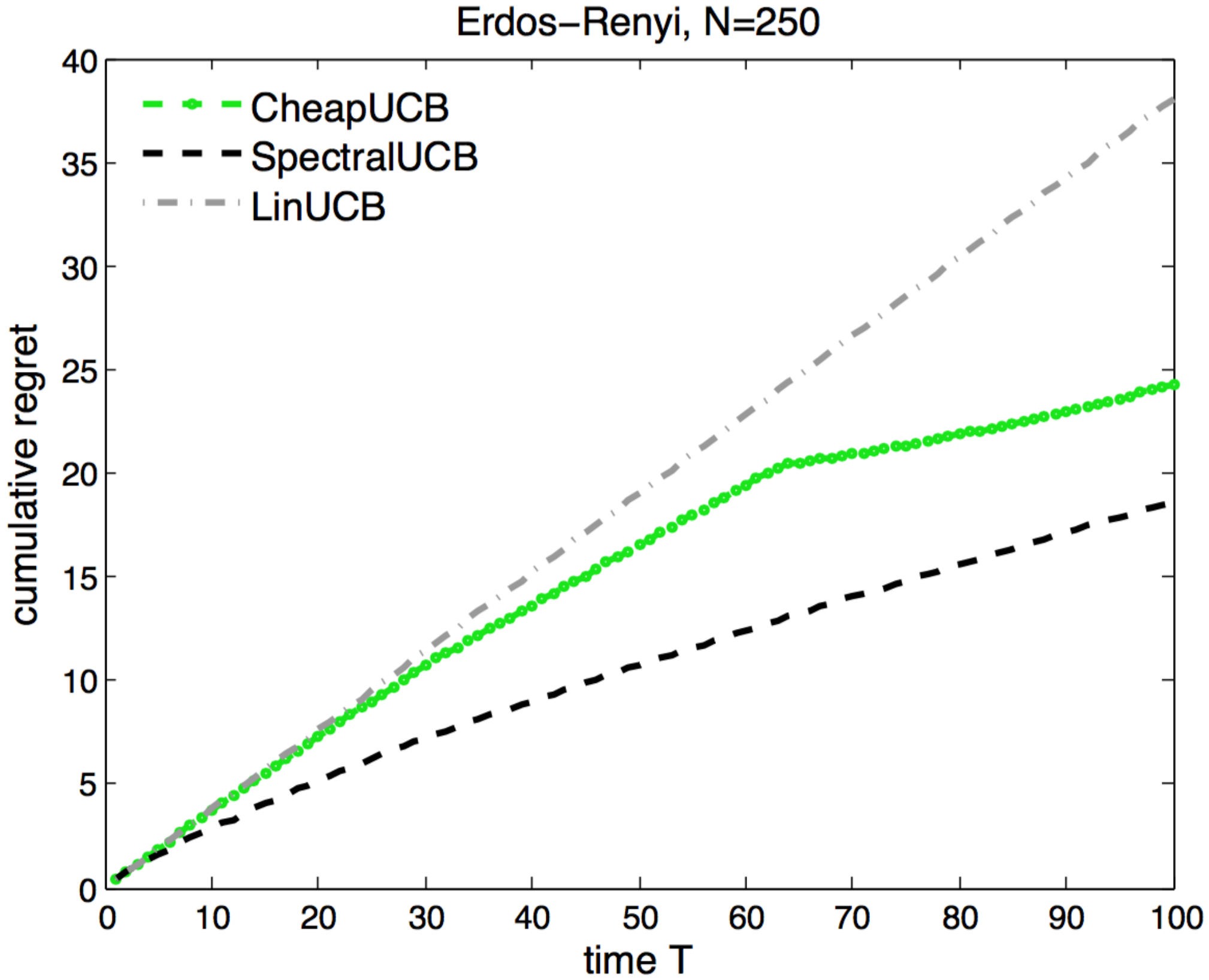}}
    	\subfigure[Cost for ER graph ]{\includegraphics[scale=0.2]{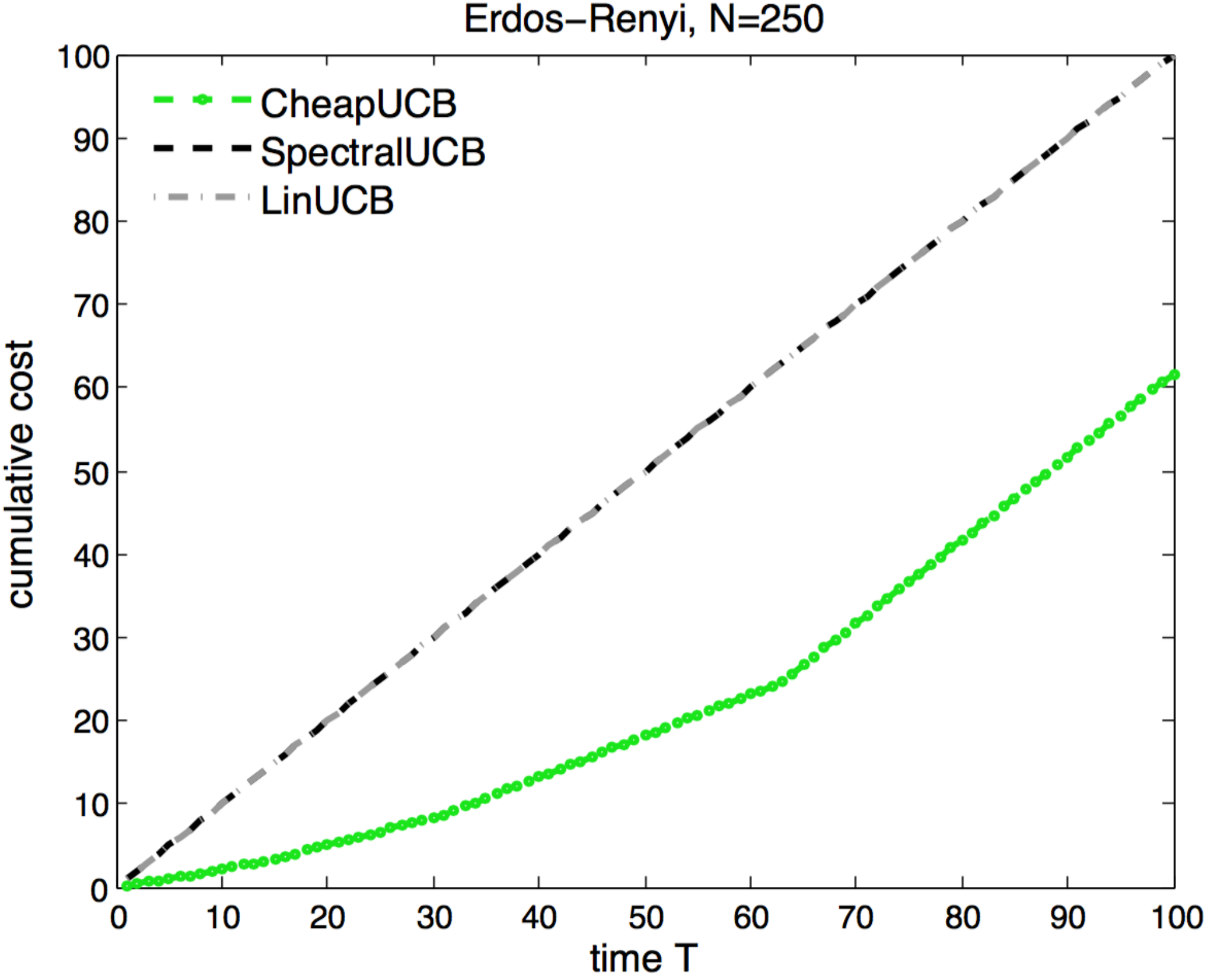}}
    	\caption{Regret and Cost for Barab\' asi-Albert (BA) and Erd\H os-R\'enyi (ER) graphs with N=250 nodes and $T=100$}
    	\label{fig:SimRandomGraphs}
    	\vspace{-.3cm}
    \end{figure*}
\subsection{Computational complexity and scalability}
The computational and scalability issues of CheapUCB are essentially those 
associated with the SpectralUCB, i.e., obtaining eigenbasis of the graph 
Laplacian, matrix inversion and computation of the UCBs. Though CheapUCB uses 
larger sets of arms or probes at each step, it needs to compute only $N$ UCBs 
as $|\mathcal{S}_w|=N$ for all $w$. The $i$-th probe in the set $\mathcal{S}_w$ can 
be computed by sorting the elements of the edge weights $W(i,:)$  and assigning 
weight $1/w$ to the first $w$ components can be done in order $N\log N$ 
computations. As \citet{valko2014spectral}, we speed up matrix inversion 
using iterative update \cite{Springer2005_TheSchurComplement_Zhang}, 
and compute the eigenbasis of symmetric Laplacian matrix using fast symmetric 
diagonally dominant solvers as CMG 
\cite{koutis2011combinatorial}. 

\vspace{-.3cm}
\section{Experiments}
\label{Sec: Experiments}
\vspace{-.1cm}
We evaluate and compare our algorithm with SpectralUCB which is shown to outperform its competitor LinUCB for learning on graphs with large number of nodes. 
To demonstrate the potential of our algorithm in a more realistic scenario we also provide experiments on Forest Cover Type dataset. We set $\delta=0.001$, $R=0.01$, and $\lambda=0.01$. 

\vspace{-.2cm}

\subsection{Random graphs models}
\label{subsec:RandomGraphs}
We generated graphs from two graph models that are widely used to analyze connectivity in 
social networks. First, we generated a {\em Erd\H os-R\'enyi} (ER) graph with 
each 
edge sampled with probability 0.05 independent of others. Second, we generated a
{\em Barab\' asi-Albert} (BA) graph with degree parameter 3. The weights of the 
edges of these graphs we assigned uniformly at random. 

To obtain a reward function $f$, we randomly generate a sparse vector 
$\alpha^*$ with a small $k \ll N$ and use it to linearly combine the 
eigenvectors of the graph Laplacian as $f=\boldsymbol{Q}\boldsymbol{\alpha}^*$, where $\boldsymbol{Q}$ is the 
orthonormal matrix derived from the eigendecomposition of the graph Laplacian. 
We ran our algorithm on each graph in the regime $T< N$. In the plots 
displayed we used $N=250$, $T=150$ and $k=5$. We averaged the experiments 
over 100 runs.

From Figure~\ref{fig:SimRandomGraphs}, we see that the 
cumulative regret performance of CheapUCB is slightly worse than for  
SpectralUCB, but significantly better than for LinUCB. However, in 
terms of the cost CheapUCB provides a gain of at least 30 \% as compared 
to both SpectralUCB and LinUCB. 

\vspace{-.2cm}
\subsection{Stochastic block models}
Community structure commonly arises in many networks. Many nodes can be naturally grouped together into a tightly knit collection of clusters with sparse connections among the different clusters. Graph representation of such networks often exhibit dense clusters with sparse connection between them. Stochastic block models are popular in modeling such community structure in many real-world networks \cite{NAS2002_CommunityStructureinSocial_GirvanNewman}.

\begin{figure*}[!t]
   	
   	\centering
   		\vspace{-.3cm}
   	\subfigure[Regret for SBM]{\includegraphics[scale=0.15]{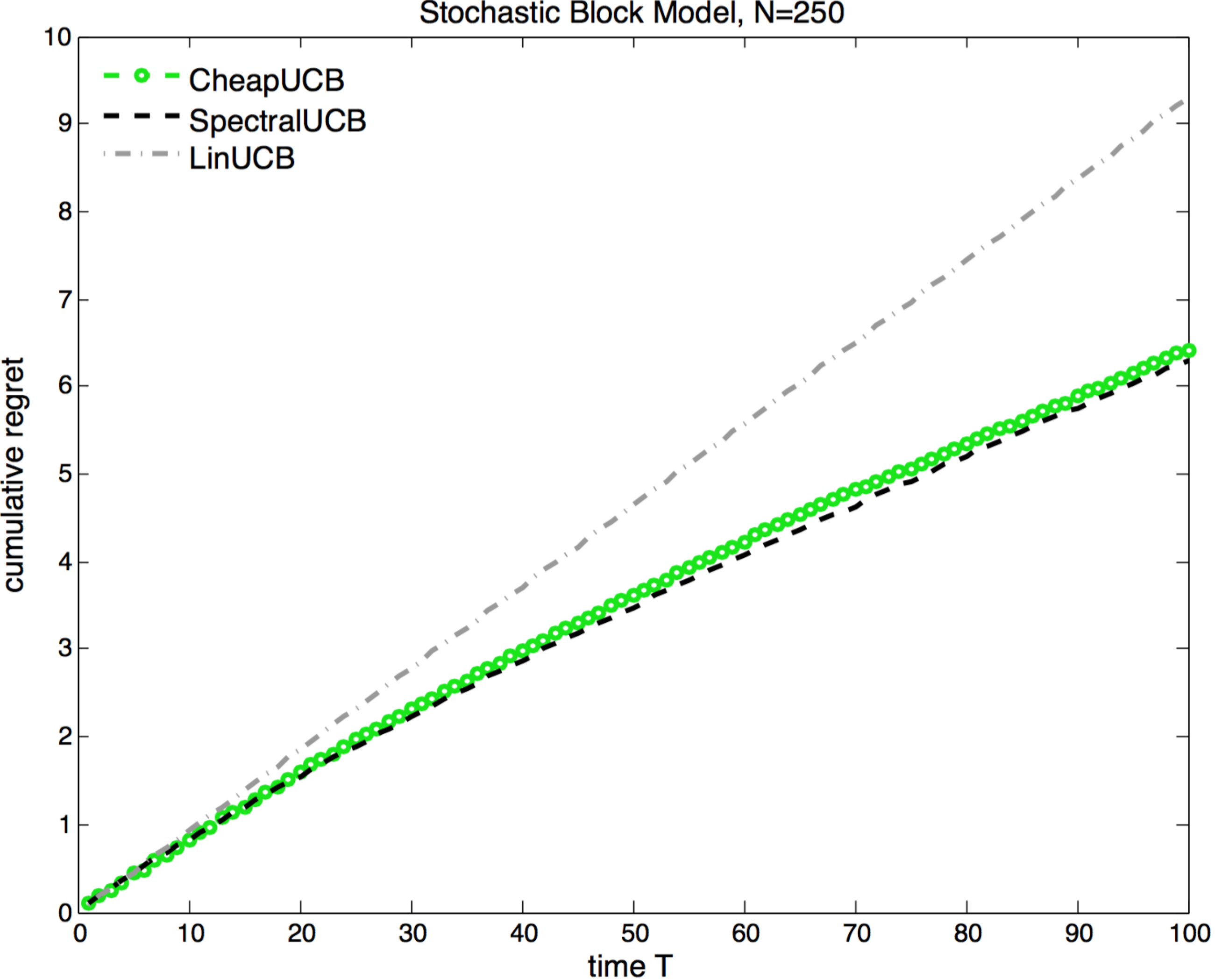}}
   	\subfigure[Cost for SBM]{\includegraphics[scale=0.15]{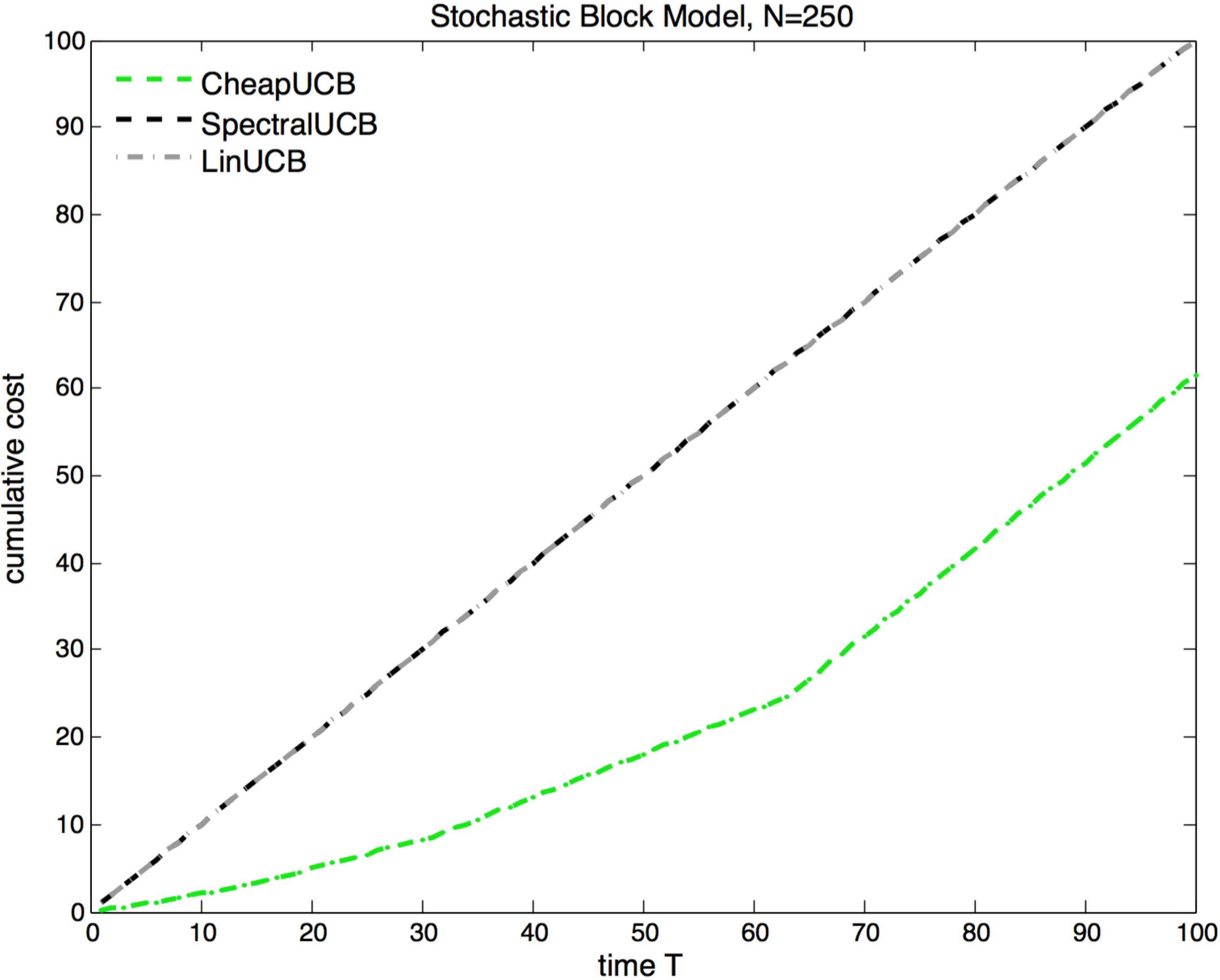}}
   	\subfigure[Regret for Forest data ]{\includegraphics[scale=0.2]{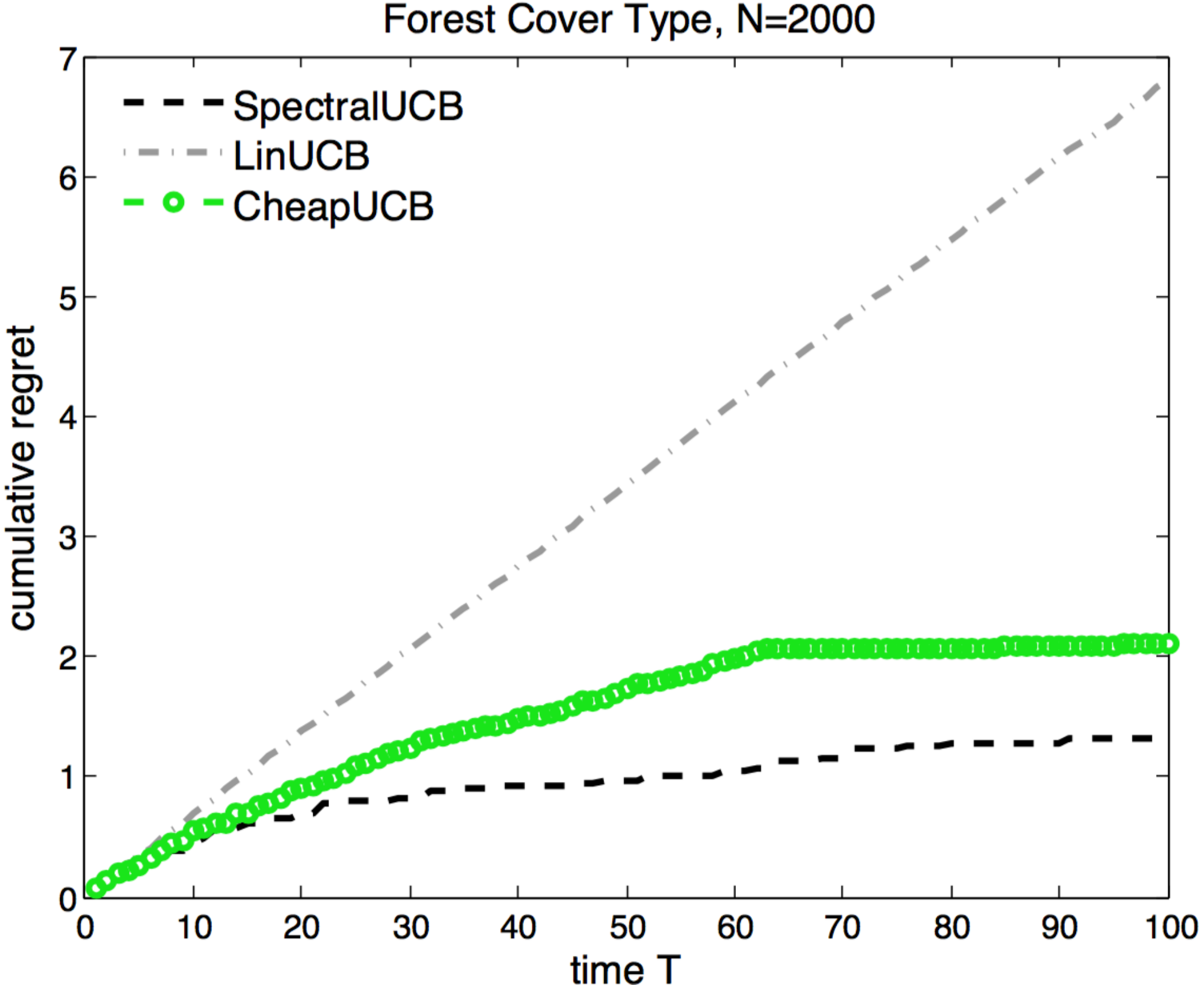}}
   	\subfigure[Cost for Forest data ]{\includegraphics[scale=0.2]{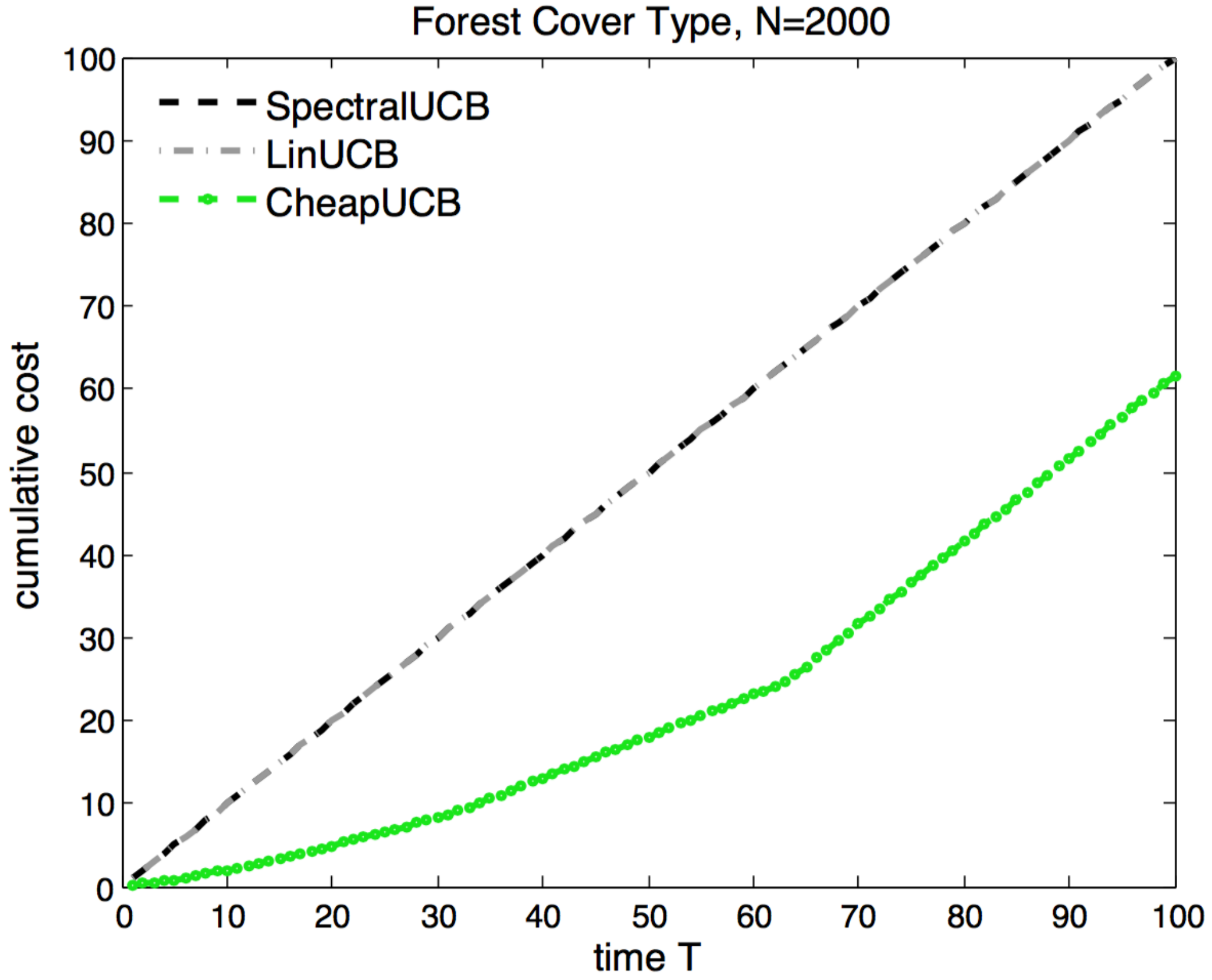}}
   	\caption{(a) Regret and (b) Cost for Stochastic block model with N=250 nodes and 4 blocks. (c) Regret and (d) Cost on the `Cottonwood' cover type of the forest data.}
   		\label{fig:SBM_FCT}
   	\vspace{-.1cm}
   \end{figure*}

The adjacency matrix of SBMs exhibits a block triangular behavior. A generative model for SBM is based on connecting nodes within each block/cluster with high probability and nodes that are in two different blocks/clusters with low probability. For our simulations, we generated an SBM as follows. We  grouped $N=250$  nodes into $4$ blocks of size $100$, $60$, $40$ and $50$, and connected nodes within each block with probability of $0.7$. The nodes from the different blocks are connected with probability $0.02$. We generated the reward function as in the previous subsection. The first $6$ eigenvalues of the graph are $0,3,4,5, 29, 29.6, \ldots$, i.e., there is a large gap between $4$th and $5$th eigenvalues, which confirms with our intuition that there should be $4$ clusters (see Prop.~2). As seen from (a) and (b) in Figure \ref{fig:SBM_FCT}, in this regime CheapUCB gives the same performance as SpectralUCB at a significantly lower cost, which confirms Theorem~2~(i) and Proposition ~2. 
%
%\begin{figure}[ht]
%	\vspace{-1.8cm}
%	\hspace{-1cm}
%	\begin{minipage}{0.2\linewidth}
%		\centering
%		\includegraphics[scale=.23]{../Simulations/SBMRegret.pdf}
%		\label{fig:SBMRegret}
%	\end{minipage}
%	\hspace{2.5cm}
%	\begin{minipage}{0.1\linewidth}
%		\centering
%		\includegraphics[scale=.23]{../Simulations/SBMCost.pdf}
%		\label{fig:SBMCost}
%	\end{minipage}
%	\vspace{-2cm}
%	\caption{Regret \& Cost for Stochastic Block Model Graph}
%	\label{fig:SBMGraph}
%\end{figure}

%\vspace{-.2cm}
\subsection{Forest Cover Type data}
\label{subsec:ForestCoverType}

As our motivation for cheap bandits comes 
from the scenario involving sensing costs, we performed experiments
on the \textit{Forest Cover Type} data, a collection of $581021$ labeled samples each 
providing observations on $30m \times 30m$ region of a forest area. This dataset was chosen to match the radar motivation from the introduction, namely, we can view sensing the forest area from above, when vague sensing is cheap and specific sensing on low altitudes is costly. This dataset was already used to evaluate
a bandit setting by~\citet{NIPS10_ParameticBandits_FilipCappGarviSzep}.

The labels in Forest Cover Type data indicate the dominant species of trees (cover type) in a given region region. The observations are 12 `cartographic' measures of the regions and are used as 
independent variables to derive the cover types. Ten of the 
cartographic measures are quantitative and indicate the distance of the regions  with respect to some reference points. The other two are qualitative binary variables indicating presence of certain characteristics. 

In a forest area, the cover type of a region depends on the geographical 
conditions which mostly remain similar in the neighboring regions.  Thus, the 
cover types change smoothly over the neighboring regions and likely to be concentrated in some parts of forest. Our goal is to find 
the region where a particular cover type has the highest concentrated.
For example, such requirement arises in aerial reconnaissance, where an air 
borne vehicle (like UAV) collects ground information through a series of measurements  to identify the regions of interests. In such applications, 
larger areas can be sensed at higher altitudes more quickly (lower cost) but 
this sensing suffers a lower resolution. On the other hand, smaller areas can 
be sensed at lower altitudes but at much higher costs.

To find the regions of high concentration of a given cover type, we first 
clustered the samples using only the quantitative attributes ignoring all the 
qualitative measurements as done in  \cite{NIPS10_ParameticBandits_FilipCappGarviSzep}. We generated $2000$ clusters (after normalizing the data to lie in the intervals $[0\; 1]$) using $k$-means 
with Euclidean distance as a distance metric. For each cover type, we  defined reward on clusters as the fraction of samples in the cluster that have the 
given cover type. We then generated graphs taking cluster centers as nodes and 
connected them with edge weight $1$ that have similar rewards using 10 nearest-neighbors method.  Note that neighboring clusters are geographically closer and will have similar cover types 
making their rewards similar. 

\begin{figure}[!h]
 \centering
\includegraphics[scale=.3]{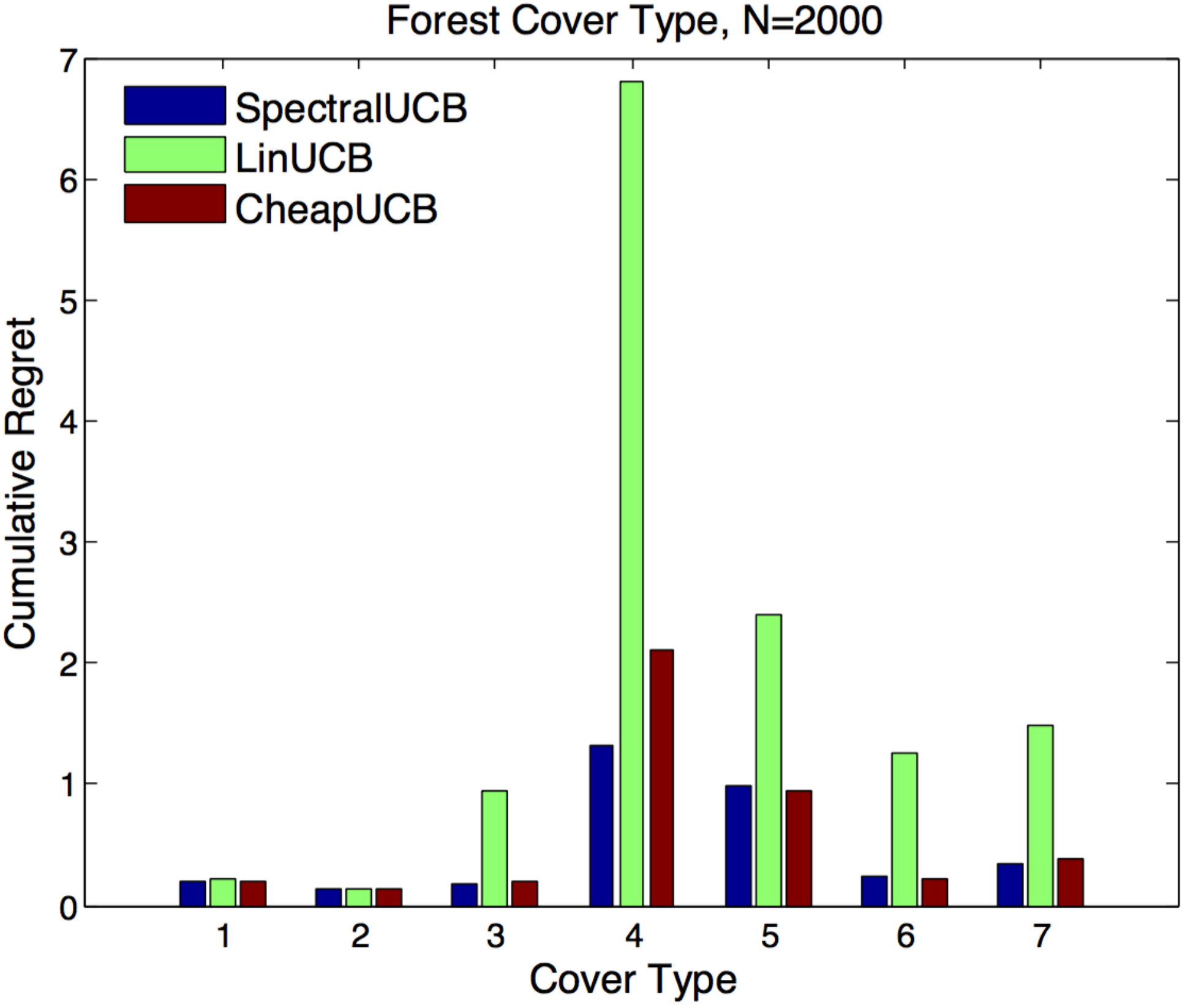}
\caption{\small Cumulative regret for different cover types of the forest cover type data set with 2000 clusters: 1- Spruce/Fir, 2- Lodgepole Pine, 3- Ponderosa Pine, 4- Cottonwood/Willow, 5- Aspen, 6- Douglas-fir, 7- Krummholz.}
\label{fig:FCT_All}
\end{figure}

We first considered the `Cottonwood/Willow' cover type for which nodes' rewards 
varies from 0 to 0.068. We plot the cumulative regret and cost in 
(c) and (d) in Figure \ref{fig:SBM_FCT} for $T=100$. As we can see, the cumulative regret of the 
CheapUCB saturates faster than LinUCB and its performance is similar to that 
of SpectralUCB. And compared to both LinUCB and SpectralUCB total cost of 
CheapUCB is less by 35 \%. We also considered reward functions for all the 7 
cover types and the cumulative regret is shown in Figure~\ref{fig:FCT_All}. 
Again, the cumulative regret of CheapUCB is smaller than LinUCB and close to 
that of SpectralUCB with the cost gain same as in  Figure~\ref{fig:SBM_FCT}(d) for 
all the cover types.

%\begin{figure}[!h]
%\vspace{-2.3cm}
%\hspace{-1cm}
% \begin{minipage}{0.2\linewidth}
%  \centering
%   \includegraphics[scale=.3]{../Simulations/FCT4Regret.pdf}
%   \label{fig:FCTRegret}
% \end{minipage}
% \hspace{2cm}
% \begin{minipage}{0.2\linewidth}
%   \centering
%   \includegraphics[scale=.3]{../Simulations/FCT4Cost.pdf}
%   \label{fig:FCTCost}
% \end{minipage}
%  \vspace{-3cm}
%\caption{\small Forest Cover Type: Cumulative regret and cumlative cost for reward function defined for the `Cottonwood' cover type.}
%\label{fig:FCT}
%\end{figure}

\section{Conclusion}
\label{Sec: Conclusion}
We introduced \textit{cheap bandits}, a new setting that aims to minimize  
sensing cost of the group actions while attaining the state-of-the-art regret guarantees in terms of effective dimension. The main advantage over typical bandit 
settings is that it models situations where getting the average reward from 
a set of neighboring actions is less costly than getting a reward from a single 
one. For the stochastic rewards, we proposed and evaluated CheapUCB, an 
algorithm that guarantees a cost gain linear in time. In future, we plan to extend 
this new sensing setting to other settings with limited feedback, such as 
contextual, combinatorial and non-stochastic bandits. As a by-product of our analysis, we establish a $\Omega(\sqrt{dT})$ lower bound on the cumulative regret for a class of graphs with effective dimension $d$.

\section*{Acknowledgment}
This material is based upon work partially supported by NSF Grants CNS-1330008, CIF-1320566, CIF-1218992, and the U.S.~Department of Homeland Security, Science and Technology Directorate, Office of University Programs, under Grant Award 2013-ST-061-ED0001. The views and conclusions contained in this document are those of the authors and should not be interpreted as necessarily representing the official policies, either expressed or implied, of the U.S.~Department of Homeland Security or the National Science Foundation.
%This work was also supported by the French Ministry of Higher Education and Research and the French National Research Agency (ANR) under project ExTra-Learn n.ANR-14-CE24-0010-01.
%\appendix
\bibliography{library,Graph}
\bibliographystyle{icml2015}

\section{Proof of Proposition \ref{prop: LowerBound}}
For a given policy $\pi, \BS{\alpha}^*,T$, and a graph $G$ define expected cumulative reward as \[Regret(T, \pi,\BS {\alpha}^*,G)=\mathbb{E}\left [ \sum_{t=1}^T \tilde{\BS{s}}_*\BS{\alpha}_* -   \tilde{\BS{s}}_t\BS {\alpha}^* | \U \right ]\]
where $\tilde{\BS{s}_t}=\pi^\prime(t)\BS{Q}$, and $\BS{Q}$ is the orthonormal basis matrix corresponding to Laplacian of $G$. 
Let $\mathcal{G}_d$ denote the family of graphs with effective dimension $d$. 
Define $T$- period risk of the policy $\pi$ 
\[Risk(T, \pi)=\max _{G \in \mathcal{G}_d}\displaystyle \max_{\begin{subarray}{l} \U \in \mathcal{R}^N\\
	\norm{\U}{\Lambda}<c\end{subarray}}\left [ Regret(T, \pi, \U, G)\right ] \]

We first establish that their exists a graph with effective dimension $d$, and a class of smooth reward functions defined over it with parameters $\U$'s in a $d$-dimensional vector space.

\begin{lemma}
	Given $T$, there exists a graph $\hat{G} \in \mathcal{G}_d$ such that 
	\[\displaystyle \max_{\begin{subarray}{l} \U \in \mathcal{R}^d\\
			\norm{\U}{\Lambda}<c\end{subarray}}\left [ Regret(T, \pi, \U, \hat{G})\right ] \leq  Risk(T, \pi)\]
	\end{lemma}
{\bf Proof:}
   We prove the lemma by explicit construction of a graph. Consider a graph $G $ consisting of $d$ disjoint connected subgraphs denoted as $G_j: j=1,2\ldots,d$. Let the nodes in each subgraph have the same reward. The set of eigenvalues of the graph are $\{0, \hat{\lambda}_1,\cdots,\hat{\lambda}_{N-d} \}$, where eigenvalue $0$ is repeated $d$ times. Note that the set of eigenvalues of the graph is the union of the set of eigenvalues of the individual subgraphs. Without loss of generality, assume that $\hat{\lambda_1} > T/d\log(T/\lambda +1 )$ (this is always possible, for example if subgraphs are cliques). Then, the effective dimension of the graph $G$ is $d$.  
	Since the graph separates into $d$ disjoint subgraphs,  we can split the reward function $\BS{f}_\alpha=\BS{Q\alpha}$ into $d$ parts, one corresponding to each subgraph. We write $\BS{f}_j=\BS{Q_j}\BS{\alpha}_j$ for $j=1,2,\ldots, d$, where $\BS{f}_i$ is the reward function associated with $G_j$, $\BS{Q}_j$ is the orthonormal matrix corresponding to Laplacian of $G_j$, and $\BS{\alpha}_i$ is a sub-vector of $\BS{\alpha}$ corresponding to $G_j$. 
	
	Write $\BS{\alpha}_j=\BS{Q}_j ^\prime \BS{f}_j$. Since $\BS{f}_j$ is a constant vector and, except for one , all the columns in $
	\BS{Q}_j$ are orthogonal to  $\BS{f}_j$, it is clear that $\BS{\alpha}_j$ has only one non-zero component. We conclude that for the reward functions that is constant on each subgraphs $\BS{\alpha}$ has only $d$ non-zero components and lies $d$-dimensional space. The proof is complete by setting $\hat{G}=G$

     Note that a graph with effective dimension $d$ cannot have more than $d$ disjoint connected subgraphs. 
     Next, we restrict our attention to graph $\widehat{G}$ and rewards that are piecewise constant on each clique. That means that the nodes in each clique have the same reward. Recall that action set $\mathcal{S}_D$ consists of actions that can probe a node or a group of neighboring nodes. Therefore, any group action will only allow us to observe average reward from a group of nodes within a clique but not across the cliques. Then, all node and group actions used to observe reward from within a clique are indistinguishable. Hence, the $\mathcal{S}_D$ collapses to set of $d$ distinct actions one associated with each clique, and the problem reduces to that of selecting a clique with the highest reward. We henceforth treat  each clique as an arm where all the nodes within the same clique share the same reward value.

     %      represented by $\BS{\alpha}^* \in \mathcal{R}^d$. Define $T$-period Bayes risk over the space $\mathcal{R}^d$ as follows:
     %      
     We now provide a lower bound on the expected regret defined as follows
     \begin{equation}
     \label{eqn:AdverRisk}
     \widetilde{Risk}(T,\pi, \widehat{G})=\mathbb{E} \left[Regret\left(T,\pi,\U, \widehat{G}\right)\right], 
     \end{equation}
     where expectation is over the reward function on the arms.

     To lower bound the regret we follow the argument of \citet{auer2002nonstochastic} and their Theorem 5.1, where an adversarial setting is considered and the expectation in (\ref{eqn:AdverRisk}) is over the reward functions generated randomly according to Bernoulli distributions. We generalize this construction to our case with Gaussian noise. The reward generation process is as follows: 
     
     Without loss of generality choose cluster $1$ to be the good cluster. At each time step $t$, sample reward of cluster $1$ from the  Gaussian distribution with mean $\frac12 + \xi$ and unit variance. For all other clusters, sample reward from the Gaussian distribution with mean $\frac12$ and unit variance.
     
     The rest of the proof of the arguments follows exactly as in the proof of Theorem 5.1\cite{auer2002nonstochastic} except at their Equation 29. To obtain an equivalent version for Gaussian rewards, we use the relationship between the $L_1$ distance of Gaussian distributions and their KL divergence. We then apply the formula for the KL divergence between the Gaussian random variables to obtain equivalent version of their Equation 30. Now note that, $\log(1-\xi^2) \sim - \xi^2$  (within a constant). Then the proof follows silmilarly by setting $\xi=\sqrt{d/T}$ and noting that the $L_2$ norm of the mean rewards is bounded by $c$ for an appropriate choice of $\lambda$.

\section{Proof of Proposition \ref{prop:Smoothness2}}

In the following, we first we give some definitions and related results.

\begin{defn}[k-way expansion constant \cite{STOC12_MultiwaySpectralPartioning_LeeGharanTrevisan}]
	Consider a graph $G$ and $\mathcal{X} \subset \mathcal{V}$ let
	\[\phi_{G}(\mathcal{X}):=\phi(\mathcal{X}) = \frac{|\partial \mathcal{X}|}{V(\mathcal{X})},\]
	where $V(\mathcal{X})$ denote the sum of the degree of nodes in $\mathcal{X}$ and $|\partial \mathcal{X}|$ denote the number of edges between the nodes in $\mathcal{X}$ and $\mathcal{V}\backslash \mathcal{X}$.\\
	For all $k > 0$, $k-$way expansion constant is defined as 
	\[\rho_\mathcal{G} (k)=\min \left \{ \max \phi(\mathcal{V}^i): \cap_{i=1}^k \mathcal{V}^i= \varnothing, |\mathcal{V}^i|\neq 0 \right \}.\]
\end{defn}
Let $\mu_1\leq \mu_2,\dots, \leq \mu_N$ denote the eigenvalues of the normalized Laplacian  of $G$. 
\begin{thm}[\cite{SODA14_PartioningIntoExpander_GharanTrevisan}]
	\label{thm:ConductanceBounds}
	Let $\varepsilon >0 $ and $\rho(k+1)>(1+\varepsilon)\rho(k)$ holds for some $k>0$. Then the following holds:
	\begin{equation}
	\label{eqn:ExpanderConstBounds}
	\mu_k/2 \leq \rho(k)\leq \mathcal{O}(k^2)\sqrt{\mu_k}
	\end{equation}
	There exits a $k$ partitions $\{\mathcal{V}^i:i=1,2,\cdots, k\}$ of $\mathcal{V}$ such that $forall \; i=1,2,\cdots k$		
	
	\begin{eqnarray}
	\label{eqn: OuterConductance}
	\phi(\mathcal{V}^i)&\leq& k\rho(k) \quad \text{and}\\
	\label{eqn:InnerConductance}
	\phi(G[\mathcal{V}^i])&\geq& \varepsilon\rho(k+1)/14k
	\end{eqnarray}
	where $\phi(G[\mathcal{X}])$ denotes the Cheeger's constant (conduntance) of the subgraph induced by $\mathcal{X}$.
\end{thm}

\begin{defn}[Isoperimetric number]
	\[\theta(G)=\left\{\min \frac{\partial \mathcal{X}}{|\mathcal{X}|}: |\mathcal{X}|\leq \mathcal{X}/2\right\}.\]
\end{defn}

Let $\lambda_1\leq \lambda_2,\ldots, \leq \lambda_N$ denotes the eigenvalues of the unnormalized Lapalcian of $G$.The following is a standard result.

\begin{equation}
\label{eqn:Isoperimetric}
\lambda_2/2 \leq \theta(G) \leq \sqrt{2\D \lambda_2}.
\end{equation}

{\bf Proof:}
The relation $\lambda_{k+1}/\lambda_{k}\geq \mathcal{O}(k^2)$ implies that $\mu_{k+1}/\mu_{k}\geq \mathcal{O}(k^2)$. Using the upper and lower bounds on the eigenvalues in  (\ref{eqn:ExpanderConstBounds}), the relation 
$\rho_{k+1} \geq (1+\varepsilon)\rho_k$ holds for some $\varepsilon>1/2$.  Then, applying Theorem \ref{thm:ConductanceBounds} we get $k$-partitions satisfying (\ref{eqn: OuterConductance})-(\ref{eqn:InnerConductance}).  
Let $\BS{L}_i$ denote the Laplacian induced by the subgraph $G[\mathcal{V}^j]=(\mathcal{V}^j, \mathcal{E}^j)$ for $j=1,2,\cdots k$. By the quadratic property of the graph Laplacian we have
	
\begin{eqnarray}
\lefteqn{\boldsymbol{f^\prime L f}=\sum_{(u,v) \in \mathcal{E}} (f_u-f_v)^2 }\\
	&=& \sum_{j=1}^k\sum_{(u,v) \in \mathcal{E}_j} (f_u-f_v)^2 \\
	&=& \sum_{j=1}^k  \boldsymbol{f_j^\prime L_j f_j} 
	\end{eqnarray}
	where $\BS{f}_j$ denote the reward vector on the induced subgraph $G_j:=G[\mathcal{V}^j]$ In the following we just focus on the optimal node. The same arguments holds for any other node. 
	Without loss of generality assume that the node with optimal reward lies in subgraph $G_l$ for some $1\leq l \leq d$. From the last relation we have $\BS{f}_l^\prime \BS{l}_l \BS{f}_l \leq c$. The reward functions on the subgraph $G_l$ can be represented as $\BS{f}_l=\BS{Q}_l\BS{\alpha}_l$ for some $\BS{\alpha}_l$, where $\BS{Q}_l$ satisfies $\BS{L}_i=\BS{Q}_l^\prime\BS{\Lambda}_L\BS{Q}_l$ and $\BS{\Lambda}_l$ denotes the diagonal matrix with eigenvalues of $\BS{\Lambda}_l$.   
	We have
	\begin{eqnarray*}
		\lefteqn{|F_G(\boldsymbol{s}_*)-F_{G}((\boldsymbol{s}_*^w)|=| F_{G_l}(\boldsymbol{s}_*)-F_{G_l}(\boldsymbol{s}_*^w) | }\\
		&\leq& \|\boldsymbol{s}_* -\boldsymbol{s}_*^w \|\|\BS{Q}_l \BS{\alpha}_l\| \\
		&\leq& \left(1-\frac{1}{w}\right)\| \BS{Q}_l  \BS{\Lambda}_l^{-1/2}\| \|\BS{\Lambda}_l^{1/2}\BS{\alpha}_l \|\\
		&\leq& \frac{c}{\sqrt{\lambda_2(G_l)}} \quad \text{From Chauchy-Schwarz}\\
		&\leq &  \frac{\sqrt{2\D}c}{{\theta(G_l)}}  \quad \;\;\text{From (\ref{eqn:Isoperimetric})}\\
				&\leq &  \frac{\sqrt{2\D}c}{{\phi(G_l)}}  \quad \;\;\text{Using $\theta(G_l)\geq \phi(G_l)$}\\
		&\leq & \frac{14k\sqrt{2\D}c}{\varepsilon\rho(k+1)} \quad \text{From Th.1, Eq. (\ref{eqn:InnerConductance})} \\
		&\leq & \frac{56k\sqrt{2\D}c}{\mu_{k+1}} \quad \text{From Th.1, Eq. (\ref{eqn:ExpanderConstBounds})}\\
				&\leq & \frac{56k\D\sqrt{2\D}c}{\lambda_{k+1}} \quad \text{ Using $\mu_{k+1}\geq \lambda_{k+1}/\D$}.
	\end{eqnarray*}
This completes the proof.

\section{Analysis of CheapUCB}

For a given confidence parameter $\delta$
define \[\beta=2R \sqrt{d\log \left (1+\frac{T}{\lambda}\right)+ 2 \log \frac{1}{\delta}} + c,\]
and consider the ellipsoid around the estimate $\boldsymbol{\hat\alpha}_t$
\[ C_t=\{\boldsymbol{\alpha}: \| \boldsymbol{\hat{\alpha}}_t 
-\boldsymbol{\alpha} \|_{V_t}\leq \beta\}.\]

We first state the following results from 
\cite{NIPS11_ImprovedAlgortihms_YadkoPalSzepes}, 
\cite{COLT08_StochasticLinearOptimization_DaniHayesKakad}, and 
\cite{valko2014spectral}
\begin{lemma}[Self-Normalized Bound]
	\label{lma:SelfNormalized}
	Let $\boldsymbol{\xi}_t= \sum_{i=1}^{t}\tilde{\mathbf{s}}_i\varepsilon_i$ and 
	$\lambda > 0$. Then, for any $\delta > 0$, with probability at least $1-\delta$ 
	and for all $t >0$,  \[\norm{\xi_t}{V_t^{-1}} \leq 
	\beta.\]
\end{lemma}
\begin{lemma}
	\label{lma:UpperBoundMatrix}
	Let $V_0=\lambda I$. We have:
	\begin{eqnarray*}
		\log \frac{\det(V_t)}{\det(\lambda I)} &\leq& \sum_{i=1}^{t} \norm{\GFT_i}{V_{i-1}^{-1}} \| 
		\tilde{\mathbf{s}}_i \| _{V_{i-1}^{-1}} \leq 2 \log 
		\frac{\det(V_{t+1})}{\det(\lambda I)} \\.
	\end{eqnarray*}
\end{lemma}

\begin{lemma}
	\label{lma:ConfidenceEllipsoid}
	Let $\| \boldsymbol{\alpha}^* \|_2 \leq c$. Then, with probability at least 
	$1-\delta$, for all $t \geq 0$ and for any $\mathbf{x}\in \mathcal{R}^n$ we have 
	$\boldsymbol{\alpha}^* \in C_t$ and
	\[|\mathbf{x}\cdot (\boldsymbol{\hat\alpha}_t - \boldsymbol{\alpha}^*)| \leq \| \mathbf{x}\|_{\bf V_t^ 
		{-1}}\beta.\] 
\end{lemma}

\begin{lemma}
	\label{lma:EffectiveDim}
	Let $d$ be the effective dimension and $T$ be the time horizon of the algorithm. Then,
	\[\log \frac{\det( V_{T+1})}{\det(\Lambda)}\leq 2d\log \left (1+ 
	\frac{T}{\lambda}\right ).\]
\end{lemma}

\subsection{Proof of Theorem \ref{thm:Regret1}}
We first prove the case where degree of each node is at least $\log T$. 

Consider step $t \in [2^{j-1}, 2^j-1]$ in stage $j=1,2,\cdots J-1$. Recall that in this step a probe of width $J-j+1$ is selected.  Write $w_j:=J-j+1$, and denote the probe of width $J-j+1$ associated with the optimal probe $\mathbf{s}_*$ as simply $\mathbf{s}^{w_j}_*$ and the corresponding GFT as $\tilde{\mathbf{s}}^{w_j}_*$. The probe selected at time $t$ is denoted as $\mathbf{s}_t$. Note that both 
$\mathbf{s}_t$ and $\mathbf{s}^{w_j}_*$ lie in the set $\mathcal{S}_{J-j+1}$. 
For notational convenience let us denote 
\begin{equation*}
h(j):=
\begin{cases}
c^\prime \sqrt{T} (J-j+1)/\lambda_{d+1} \;\;\mbox{when (\ref{eqn:LocalSmoothness2}) holds}\\
c^\prime d /\lambda_{d+1} \;\;\mbox{when (\ref{eqn:LocalSmoothness1}) holds}.
\end{cases}
\end{equation*}

The instantaneous regret in step $t$ is

\begin{eqnarray*}
	\lefteqn{r_t
		= \tilde{\mathbf{s}}_*\cdot \boldsymbol{\alpha}^* -\tilde{\mathbf{s}}_t \cdot \boldsymbol{\alpha}^*}\\  
	&\leq & \tilde{\mathbf{s}}^{w_j}_* \cdot \boldsymbol{\alpha}^* +h(j)-\tilde{\mathbf{s}}_t \cdot \boldsymbol{\alpha}^*\\
	%&=& \tilde{\mathbf{s}}_j^\prime (\alpha^*-\hat\alpha_t) +\tilde{\mathbf{s}}_j^\prime\hat\alpha_t  -\tilde{\mathbf{s}}_t^\prime \alpha^* + ch(j)\quad \text{add and substract}\;\; \tilde{\mathbf{s}}_j^\prime\hat\alpha_t\\
	&=& \tilde{\mathbf{s}}^{w_j}_* \cdot (\boldsymbol{\alpha}^*-\boldsymbol{\hat\alpha}_t) 
	+\tilde{\mathbf{s}}^j_* \cdot \boldsymbol{\hat\alpha}_t + 
	\beta\|\tilde{\textbf{s}}^{{w_j}}_*\|_{\bf V_t^{-1}} \\
	&& - \beta\| \tilde{\textbf{s}}^{w_j}_* \| _{\bf V_t^{-1}}  
	-\tilde{\mathbf{s}}_t \cdot \U + h(j)\\
	%&& \quad \text{add and substract}\;\; \beta_t\| \tilde{\textbf{s}}_* \| 
	%_{V_t^{-1}}\\
	&\leq& \tilde{\mathbf{s}}^{w_j}_* \cdot (\boldsymbol{\alpha}^*-\boldsymbol{\hat\alpha}_t) 
	+\tilde{\mathbf{s}}_t \cdot \boldsymbol{\hat\alpha}_t + \beta\| \tilde{\textbf{s}}_t 
	\| _{\bf V_t^{-1}}\\
	&& - \beta\| \tilde{\textbf{s}}^{w_j}_* \| _{\bf V_t^{-1}}  
	-\tilde{\mathbf{s}}_t \cdot \boldsymbol{\alpha}^*+ h(j)\\
	&=& \tilde{\mathbf{s}}^{w_j}_* \cdot (\boldsymbol{\alpha}^*-\boldsymbol{\hat\alpha}_t) 
	+\tilde{\mathbf{s}}_t \cdot(\boldsymbol{\hat\alpha}_t-\boldsymbol{\alpha}^*) + \beta\| 
	\tilde{\textbf{s}}_t \| _{V_t^{-1}}\\
	&& - \beta\| \tilde{\textbf{s}}^{w_j}_* \| _{\bf V_t^{-1}}+ h(j)\\
	&\leq& \beta\| \tilde{\textbf{s}}^{w_j}_* \| _{\bf V_t^{-1}} +\beta\| 
	\tilde{\textbf{s}}_t \| _{\bf V_t^{-1}} + \beta\| 
	\tilde{\textbf{s}}_t \| _{\bf V_t^{-1}}\\
	&& - \beta\| \tilde{\textbf{s}}^{w_j}_* \| _{\bf V_t^{-1}}+ h(j)\\
	&=& 2\beta\| \tilde{\textbf{s}}_t \| _{\bf V_t^{-1}}+ h(j).
\end{eqnarray*}
We used (\ref{eqn:LocalSmoothness1})/(\ref{eqn:LocalSmoothness2}) in the first inequality. The second inequality follows from the algorithm design and the third inequality follows from Lemma  \ref{lma:ConfidenceEllipsoid}. Now, the cumulative regret of the algorithm is given by
\begin{eqnarray*}
	\lefteqn {R_T}\\
	&\leq& \sum_{j=1}^{J}\sum_{t=2^{j-1}}^{2^j-1}\min\{2,2\beta\| 
	\tilde{\textbf{s}}_t \| _{V_t^{-1}}+ h(j)\} \\
	&\leq& \sum_{j=1}^{J}\sum_{t=2^{j-1}}^{2^j-1}\min\{2,2\beta_t\| 
	\tilde{\textbf{s}}_t \| _{V_t^{-1}}\} + 
	\sum_{j=1}^{J-1}\sum_{t=2^{j-1}}^{2^j-1} h(j)\\
	&\leq& \sum_{t=1}^{T}\min\{2,2\beta_t\| \tilde{\textbf{s}}_t \| 
	_{V_t^{-1}}\} + \sum_{j=1}^{J-1}h(j)2^{j-1}.
\end{eqnarray*}

Note that the summation in the second term includes only the first $J-1$ stages. In the last stage $J$, we use probes of width $1$ and hence we do not need to use (\ref{eqn:LocalSmoothness1})/(\ref{eqn:LocalSmoothness2}) in bounding the instantaneous regret. Next, we bound each term in the regret separately.

To bound the first term  we use the same steps as in the proof of Theorem 1 \cite{valko2014spectral}. We repeat the steps below.

\begin{eqnarray}
\lefteqn{\sum_{t=1}^{T}\min\{2,2 \beta\| \tilde{\textbf{s}}_t \| 
	_{V_t^{-1}}\}}\nonumber\\
&\leq & (2+2\beta)\sum_{t=1}^{T}\min\{1,\| \tilde{\textbf{s}}_t \| 
_{\bf V_t^{-1}}\}\nonumber\\
&\leq&(2+2\beta)\sqrt{T\sum_{t=1}^{T}\min\{1, \beta_t\| \tilde{\textbf{s}}_t 
	\| _{\bf V_t^{-1}}\}^2} \nonumber\nonumber\\
\label{eqn:LogDeterminantBound}
&\leq& 2 (1+\beta)\sqrt{2T \log (|\BS{V}_{T+1}|/|\BS{\Lambda|})}\\
\label{eqn:EffectiveDimBound}
&\leq& 4 (1+\beta)\sqrt{Td\log (1+ T/ \lambda)} \\
&\leq& \left (8R\sqrt{2\log \frac{1}{\delta}+ d\log\left (1+ \frac{T}{\lambda} \right )} + 4c+4\right )\nonumber\\
&&\times \sqrt{Td\log \left (1+ \frac{T}{\lambda} \right )}\nonumber.
\end{eqnarray}
We used Lemma \ref{lma:UpperBoundMatrix} and \ref{lma:EffectiveDim} in 
inequalities (\ref{eqn:LogDeterminantBound}) and (\ref{eqn:EffectiveDimBound}) 
respectively.  The final bound follows from plugging the value of $\beta$.

\subsection{For the case when (\ref{eqn:LocalSmoothness2}) holds:} 
For this case we use $h(j)=c^\prime \sqrt{T}(J-j+1)/\lambda_{d+1}.$
First observe that $2^{j-1}h(j)$ is increasing in $1\leq j \leq J-1$. We have
\begin{eqnarray*}
	\lefteqn{\sum_{j=1}^{J-1}\frac{2^{j-1}c^\prime\sqrt{T}(J-j+1)}{\lambda_{d+1}} \leq (J-1)\frac{2^{J-1}\sqrt{T}c^\prime}{\lambda_{d+1}}}\\
	&\leq& (J-1)\frac{2^{\log_2 T -1}c^\prime \sqrt{T}}{\lambda_{d+1}} \leq (J-1)\frac{c^\prime\sqrt{T}(T/2)}{(T/d \log (T/\lambda +1))} \\
	&\leq&  dc^\prime \sqrt{T/4}\log_2 (T/2) \log(T/ \lambda +1).
\end{eqnarray*}

In the second line we applied the definition of effective dimension.

\subsection{For the case when $\lambda_{d+1} /\lambda_d \geq \mathcal{O}(d^2)$} 
For the case $\lambda_{d+1} /\lambda_d \geq \mathcal{O}(d^2)$ we use $h(j)=c^\prime d/\lambda_{d+1}.$
\begin{eqnarray*}
	\lefteqn{\sum_{j=1}^{J-1}\frac{2^{j-1}c^\prime d}{\lambda_{d+1}} \leq \frac{2^{J-1}c^\prime d}{\lambda_{d+1}}}\\
	&\leq&  c^\prime d^2 \log_2 (T/2) \log(T/ \lambda +1).
\end{eqnarray*}

Now consider the case where minimum degree of the nodes is $1<a\leq \log T$. In this case we modify the algorithm to use only signals of width $a$ in the first $\log T - a +1$ stages and subsequently the signal width is reduced by one in each of the following stages. The previous analysis holds for this case and we get the same bounds on the cumulative regret and cost. When $a=1$, CheapUCB is same as the SpectralUCB, hence total cost and regret is same as that of SpectralUCB. 

To bound the total cost, note that in stage $j$ we use signals of width $J-j+1$. Also, the cost of a signal given in  (\ref{eqn:SignalCost}) can be upper bounded as $C(\mathbf{s}_i^w)\leq \frac{1}{w}$. Then, we can upper bound total cost of signals used till step $T$ as
\begin{eqnarray*}
	\lefteqn{\sum_{j=1}^{J}\frac{2^{j-1}}{J-j+1}}\\
	&\leq& \frac{1}{2}\sum_{j=1}^{J-1}2^{j-1} + \frac{T}{2}\\
	&\leq& \frac{1}{2}\left(\frac{T}{2}-1\right)+ \frac{T}{2}=\frac{3T}{4}-\frac{1}{2}.
\end{eqnarray*}

\end{document}